# Grammaticality illusion or ambiguous interpretation? Event-related potentials reveal the nature of the missing-NP effect in Mandarin centre-embedded structures

Qihang Yang[a], Caimei Yang [a*], Yu Liao[b*], Ziman Zhuang[c]

[a] *School of Foreign Languages, Soochow University, No. 333 Ganjiang East Road, Gusu District, Suzhou, Jiangsu, China, 215006. E-mails: harryyang2025@mail.gdufs.edu.cn (Qihang Yang); cmyang@suda.edu.cn (Caimei Yang)*

[b] *School of Education, Soochow University, No. 1 Wenjing Road, Suzhou, Jiangsu, China, 215123. E-mail: yliao@suda.edu.cn*

[c] *Suzhou Industrial Park Jinji Lake School, Suzhou, Jiangsu, China, 215021 . E-mail: 2155628639@qq.com*

*Corresponding authors:

1. Caimei Yang;
   telephone number: 8615716203062;
   email: cmyang@suda.edu.cn;
   postal address: School of Foreign Languages, Soochow University, No. 333 Ganjiang East Road, Gusu District, Suzhou, Jiangsu, China, 215006

2. Yu Liao:
   telephone number: 8618625227980;
   email: yliao@suda.edu.cn
   postal address: School of Education, Soochow University, No. 1 Wenjing Road, Suzhou, Jiangsu, China, 215123

Declarations of interest: none

**Funding acknowledgement**:

This work was supported by The National Social Science Fund of China [grant numbers: 23BYY170]

**Abstract**

In several languages, omitting a verb phrase (VP) in double centre-embedded structures creates a grammaticality illusion. Similar illusion also exhibited in Mandarin missing-NP double centre-embedded structures. However, there is no consensus on its very nature. Instead of treating it as grammaticality illusion, we argue that ambiguous interpretations of verbs can best account for this phenomenon in Mandarin. To further support this hypothesis, we conducted two electroencephalography (EEG) experiments on quasi double centre-embedded structures whose complexity is reduced by placing the self-embedding relative clauses into the sentence's subject position. Experiment 1 showed that similar phenomenon even exhibited in this structure, evidenced by an absence of P600 effect and a presence of N400 effect. In Experiment 2, providing *semantic cues* to reduce ambiguity dispelled this illusion, as evidenced by a P600 effect. We interpret the results under garden-path theory and propose that word-order difference may account for this cross-linguistic variation.

## 1 Introduction

Understanding what makes language unique to human is the goal of any linguistic theory. In a review, Hauser et al. (2002) first ventured that "…recursion is the only uniquely human component of the faculty of language", which sparked intense debates and discussions about this strong hypothesis. *Centre-embedded recursion*, a type of syntactic structure which has the property of self-embedding, has been the focal type of recursive structures (Corballis, 2007; Fitch, 2010; Levinson, 2014; Martins & Fitch, 2014). One classic example of recursive centre-embedded structure is "The rat the cat the dog chased killed ate the malt." (Chomsky & Miller, 1963), where one relative clause is embedded into the other, and the self-embedding relative clauses are again embedded into the main clause, therefore known as *double centre-embedded structure* (DCE structure). This type of structure serves as a prism to scrutinize the "recursion-only hypothesis (ROH)" (Pinker & Jackendoff, 2005) and has catalysed myriad of comparative studies in exploring human and non-human animals' differences, though with contrasting results (e.g., Beckers et al., 2012; Ferrigno et al., 2020; Fitch & Hauser, 2004; Gentner et al., 2006; Lakretz & Dehaene, 2021; Liao et al., 2022; Suzuki & Matsumoto, 2022)

Uniquely human or not, our ability in processing DCE structure is notably poor. Psycholinguistic research consistently indicates that both comprehension and production of it pose significant challenges to native speakers. (Blaubergs & Braine, 1974; Blumenthal, 1966; Chomsky & Miller, 1963; Church, 1980; Kimball, 1973; Marks, 1968; Miller & Isard, 1964; Stolz, 1967). And this difficulty also bears cross-linguistic similarity (German: Bach et al. (1986); Hebrew: Schlesinger (1975); Spanish: Hoover (1992); Japanese and Korean: Babyonyshev and Gibson (1999); Hagstrom and Rhee (1997); Uehara and Bradley (1996)).

Interestingly, when an English DCE structure such as (1a) omits a middle verb phrase (1b) that renders the structure ungrammatical, native English speakers showed no less preference for the ungrammatical sentences reflected on comprehensibility/acceptability judgement tests (Frazier, 1985; Gibson & Thomas, 1999), or less reading times (Frank et al., 2016, experiment 2&3; Vasishth et al., 2010, experiment 1&2), giving the impression of grammaticality illusion (Phillips et al., 2010). Similar phenomenon is also observed in French, Spanish, and Mandarin (Gimenes et al., 2009; Huang & Phillips, 2021; Pañeda & Lago, 2020), but purported to be absent in Dutch (Frank & Ernst, 2019; Frank et al., 2016) and has contrasting results in German (Bader, 2016; Engelmann & Vasishth, 2009; Häussler & Bader, 2015; Vasishth et al., 2010).

(1) a. #The patient the nurse the clinic hired sent to the doctor met Jack.
b. *The patient the nurse the clinic hired met Jack. (Frazier, 1985)[1]

This paper focuses particularly on the online processing of Mandarin DCE structures and its related missing-NP grammaticality illusion as reported by Huang and Phillips (2021). We contend that current explanations fall short in accounting for the missing-NP effect; in addition, we suggest that Huang and Phillips' *interference-and-repair analysis* involves logical problems and many redundancies. We propose that in in Mandarin, whose relative clause is head-final, it is ambiguity in interpreting verbs that induces the missing-NP effect and contributes to the processing limitation of Mandarin DCE structure; and that this ambiguity is irresolvable. To substantiate our proposition, we conducted two ERP experiments on Mandarin quasi-DCE structure, whose complexity is reduced by placing the self-embedding RCs into the sentential subject position.

The paper is structured as follows. In Section 2 we review existing accounts that address the processing limitation of DCE structure in head-initial language and then introduce challenges related to the missing-NP effect in Mandarin. In Section 3, we detail our proposal, addressing both the aforementioned phenomenon and the processing limitation of Mandarin DCE structure; We then presents two ERP experiments to support our hypothesis in Section 4 and 5. We end our discussion in Section 6 and 7.

## 2 What causes the processing limitation of DCE structure?

In general, literature on the processing limitation of DCE structure is categorized into two distinct research streams. The first posits a competence grammar that distinguishes between competence and performance, and attributes the parser's limitations of this structure to non-linguistic factors like working memory constraints (e.g.,Bever, 1970; Chomsky & Miller, 1963; Frazier, 1985; Frazier & Fodor, 1978; Gibson, 1998; Kimball, 1973; Lewis, 1996; Miller & Chomsky, 1963; Miller & Isard, 1964; Stabler, 1994; Yngve 1960; etc.); and the other assumes no such a priori and ascribes the processing limitation of DCE structure to linguistic experience and some intrinsic architectural constraints of learning and processing (Christiansen & MacDonald, 2009; Christiansen, 1999; Lakretz et al., 2020). For reasons of space, we do not attempt to include all proposals that have addressed this question, but only include those that explained both the processing limitation of DCE structure and its related grammaticality illusion. In section 2.1, we review three major theories that have attracted the most attention, namely resource-based account, interference-based account, and experience-based account. In each sub-section, we first explain how these theories account for the difficulty in processing DCE structures, followed by a discussion on how such processing challenges contribute to the grammaticality illusion. In section 2.2, we introduce Mandarin DCE structure and the challenges it poses to the existing accounts. We agree with Huang and Phillips (2021) that existing accounts fall short of explaining the missing-NP effect in Mandarin DCE structure and meanwhile argue against their analysis of this phenomenon.

### 2.1 Existing accounts for the processing limitation of DCE structure

#### *2.1.1 Resource-based account*

The resource-based account attributes the processing limitation and its related grammaticality illusion of DCE structure to the constrained working memory system. Before we take a closer look at how limitations of working memory hinders the parser to process DCE structure, let us first dig into how the parser processes sentences under this account.

[1] First observation of this phenomenon is attributed to Janet Fodor.

As there is strong evidence for an incremental parsing process, this theoretical framework posits two sub-components necessary for parsing a sentence (Gibson, 1998; Just & Carpenter, 1992). First, the parser must store words of input string and keep them as syntactic structures. Second, the newly received strings are integrated into the ongoing structure and connected to specific syntactic objects to fulfil incomplete dependencies (e.g., connecting a verb to its arguments for theta-role assignments and case-marking). According to *the Syntactic Prediction Locality Theory* (SPLT, Gibson, 1998) and its successor Dependency Locality Theory (DLT, Gibson, 2000), these two components create *storage cost* and *integration cost*, respectively. In DLT, storage cost is deemphasized than in SPLT since "many resource complexity effects can be explained using integration cost alone" (Gibson, 2000, p. 102). For brevity, we focus solely on the *integration cost* and its prediction for processing difficulty of DCE structure.

DLT outlines two sub-components for *integration cost*: i) introducing a new discourse referent incurs a cost of 1 EU (see Gibson, 2000 for a detailed defintion of "new discourse referent"); ii) a *structural integration cost* contingent on the intervening new discourse referents between the input object α and its dependent object β. Thus, the overall integration cost for a new discourse referent is (1 + n) EUs, where "n" represents the number of intervening discourse referents. Accordingly, in parsing a DCE structure (2), DLT predicts that the second verb consumes the most resources, with a projected *integration cost* of 7 EUs. This is broken down as: i) 1 EU for introducing the new discourse referent verb "killed"; ii) 2 EUs for the integration of the verb "killed" and its dependent object "the cat" in the subject position, involving 2 intervening discourse referents "dog" and "chased"; iii) 4 EUs for the integrating the empty category in the object position and its moved position in the RC complementizer, with the intervention of 4 discourse referents "cat", "dog", "chased" and "killed".

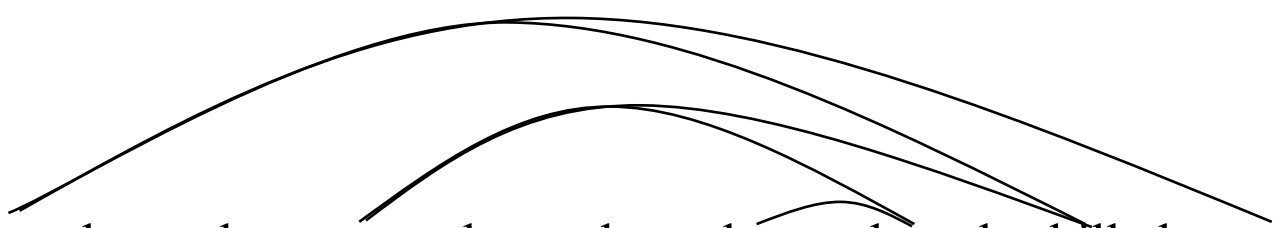

| | | | | | | | | | | | | | |
|---|---|---|---|---|---|---|---|---|---|---|---|---|---|
| (2) | The | rat | that | the | cat | that | the | dog | chased | killed | ate | the | malt |
| | 0 | 1 | 0 | 0 | 1 | 0 | 0 | 1 | 1+2 | 1+2+4 | 1+4 | 0 | 1 |

With storage cost and integration cost contributing to sentence parsing, we now come back to the question of how working memory constraints impede the parsing process of DCE structure and potentially lead to the missing-VP effect. The resource-based account assumes that there is a specific threshold for processing resources available for sentence parsing (Just & Carpenter, 1992), and exceeding this threshold significantly hampers sentence processing. Parsing DCE structure is one such case. Multi-degree of centre-embedding escalates both *storage cost* and *integration cost* to a point that available resources are inadequate for processing.

Accordingly, Gibson and Thomas (1999) based on DLT and SPLT proposed *the high memory cost pruning hypothesis* to account for the missing-VP effect in English DCE structure:

(3) "At points of high memory complexity, forget the syntactic prediction(s) associated with the most memory load." (p. 231).

DLT suggests that the middle verb incurs the highest memory load. Hence, *the high memory cost pruning hypothesis* anticipates that due to processing overload, the parser would forget the dependencies associated with the middle verb. Therefore, parsing a missing-VP DCE structure should be no more difficult than a complete DCE structure. This prediction has been substantiated through both offline and online assessments (Christiansen & MacDonald, 2009; Frank & Ernst, 2019; Frank et al., 2016; Gibson & Thomas, 1999; Gimenes et al., 2009; Vasishth et al., 2010).

*2.1.2 Interference-based account*

Different from resource-based account which emphasizes processing loads, interference-based account attributes the processing limitation of DCE structure to interference of similar linguistic objects. This account is derived from cue-based parsing models, which posit that linguistic objects are incrementally received and stored in a content-addressable memory as feature bundles or chunks, comprising syntactic and semantic information (see Lewis & Vasishth, 2005; Lewis et al., 2006; Van Dyke & McElree, 2006, 2011; Vasishth et al., 2019 for details). Processing difficulty emerges in an incremental parsing process when a sentence contains similar feature bundles awaiting to be retrieved. Example (3) gives a simplified illustration of how interference impedes sentence parsing: when the parser receives a cue from object Y that requires retrieving information from X, object Z, sharing similar features with X (category, plurality, animacy, etc.) competes for this retrieval, leading the parser to potentially misidentify Z for X. This memory interference effect is substantiated by many empirical studies (Friedmann et al., 2009; Gordon et al., 2001, 2004; Gordon et al., 2006; Humphreys et al., 2016; Jäger et al., 2017; Kaan, 2001; Liu, 2023; Schoknecht et al., 2022).

(4) X…Z…Y

Similar interference effect can also be seen in parsing DCE structure like (5). When encoded, each of the noun phrases in DCE structure predicts a verb for dependencies. Parsing $VP_1$ should cause trivial difficulty since its integration feature bundle, $NP_3$, has been recently encoded, thus benefiting from a recency effect that keeps it in the focus of attention with high activation (McElree, 2006).

| (5) | $NP_1$ | $NP_2$ | $NP_3$ | $VP_1$ | $VP_2$ | $VP_3$ |
|---|---|---|---|---|---|---|
| | The cat | the dog | the monkey | teased | chased | ran |

Encoding the second verb introduces difficulties. First, the parser must decide which attachment sites ($VP_2$ & $VP_3$) it fits into. Bader and his colleague (Bader, 2016; Häussler & Bader, 2015) suggest that the two attachment sites are similar in English; therefore, the parser is likely to integrate the second verb into site $VP_3$, resulting in processing overload upon encountering the third verb.. Second, due to the similarity of $NP_1$ and $NP_2$, as well as the parser's poor ability in serial order reference, encoding the second verb makes the retrieving (of $NP_2$) process difficult (see Lewis, 1996; Lewis & Vasishth, 2005; Lewis et al., 2006; McElree & Dosher, 1993, for details).

As to the missing-VP effect, Häussler and Bader (2015) proposed *the discrimination hypothesis*:

(6) *Discrimination hypothesis:*
"The integration of new material becomes difficult when an intervening clause separates the word that is to be integrated next from the required integration site and an incorrect but similar integration site competes for attachment." (Häussler & Bader, 2015, p. 5)

Owing to the two attachment sites' similarity, they assumed that missing-VP effect occurs when the parser mis-attaches $VP_2$ to $VP_3$ site for integration with $NP_1$. But this operation would make $NP_2$ structural "orphaned". Häussler and Bader argued that the well-studied primacy effect (e.g., Knoedler et al., 1999; Page & Norris, 1998) may explain the overlooking of this "orphanage". Since $NP_1$ is the subject of the main clause, it enjoys the primacy effect for completing its dependency with a verb. Therefore, failure to detect this orphanage occurs naturally because $NP_2$ is no longer the current focus of attention as soon as the parser completed the retrieval of $NP_3$.

*2.1.3 Experience-based account*

Experience-based theories hold that the processing difficulty of a given word is directly related to how predictable the word is given the preceding context. That is, less expected words require more processing load than those with high predictability. Similarly, statistically more frequently occurred syntactic structures are more predictable and thereby easier to process than those relatively rare structures (Futrell et al., 2020; Hale, 2001; Hale et al., 2022; Levy, 2008).

While both resource-based and interference-based accounts focus on the parser's working memory, experience-based account attributes the processing limitation of DCE structure to its rare occurrence in natural language (Karlsson, 2007). From a usage-based perspective, Christiansen and his colleagues proposed that the sole processing difficulty of DCE structure should not stem from a memory limitation, but instead be rooted in the build-up of the structure (Christiansen & MacDonald, 2009; Christiansen, 1999; Christiansen & Chater, 2015). They trained a Simple Recurrent Network (SRN) via a word-by-word prediction task; and after the training session, the model was tested on novel DCE sentences. The behaviour of their SRN model matched a human-like performance on this structure; also, it "experienced" the missing-VP effect in processing ungrammatical DCE structure. Usually enlarging the size of the model's memory unit would enhance its performance on sentence processing. However, they found that this was not the case for DCE structure. Therefore, they concluded that extrinsic memory constraints should not be the primary factor for the processing limitation of DCE structure. Although their model astonishingly predicts human performance on DCE structure, they did not explicitly provide an answer for what causes this limitation.

Cross-linguistic comparative research sheds light on this issue. Vasishth et al. (2010) found that, whereas English native speakers did show a missing-VP illusion for DCE structure, German speakers exhibited the opposite: ungrammatical DCE structure are processed more slowly than their grammatical equivalents. This outcome corroborated their prior SRN simulation, which found that missing-VP effect did not show in their model trained in a German context (Engelmann & Vasishth, 2009).

Vasishth and his colleagues (also Frank and Ernst (2019); Frank et al. (2016)) suggested that the behavioral disparities observed between English and German native speakers likely originate from differences in linguistic statistics of the two languages. German follows an SOV word order, positioning verbs in relative clauses at the sentence's end; conversely, in English, only object-extracted relative clauses are verb-final. The prevalence of verb-final clauses may increase the parser's sensitivity to the occurrence of verbs in a German context. This hypothesis gains further support from a study employing *artificial sequence learning* (ASL) paradigm, which revealed that German native speakers similarly exhibited a similar missing-VP effect in $A_nB_n$ sequence learning (de Vries et al., 2012). The aforementioned evidence suggests that linguistic statistics might be the primary factor that determines whether in a language exists missing-VP effect, as opposed to a universal memory constraint.

### 2.2 Mandarin DCE structure and the missing-NP effect

Mandarin Chinese, similar to English, employs an SVO word order; however, its relative clause structure is head-final. Examples (7a) – (8b) give a direct comparison of Mandarin and English's subject-extracted relative clause (SRC) and object-extract relative clause (ORC), where Mandarin *de* serves as relativizer. Owing to this structural disparity, constructing a DCE sentence in Mandarin necessitates embedding one SRC within another, resulting in a [V [V N *de* N]$_1$ *de* N]$_2$ sequence. By positioning this complex NP into the object position of a sentence, we can get a Mandarin DCE sequence "[$N_0$ $V_1$ [$V_2$ [$V_3$ $N_1$ *de* $N_2$] *de* $N_3$]]", exemplified in (9a), where the innermost RC "…$V_1$ $N_1$ *De* $N_2$…" is doubly centre-embedded. Omitting the second head-noun ($N_3$) and its relativizer *de* renders the DCE structure incomplete, as in (9b).

**Mandarin subject-extracted RCs:**

7a) [$_{RC}$ _$_{gap}$ bāngzhù nǎinai de] [$_{filler}$ yéyé] fēicháng shànliáng
[$_{RC}$ _$_{gap}$ helped grandma REL] [$_{filler}$ grandpa] very kind

Its English translation:

7b) [$_{filler}$ The grandpa] [$_{RC}$ who _$_{gap}$ helped the grandma] is very kind.

**Mandarin object-extracted RCs:**

8a) [$_{RC}$ yéyé bāngzhù _$_{gap}$ de] [$_{filler}$ nǎinai] fēicháng shànliáng
[$_{RC}$ grandpa helped _$_{gap}$ REL] [$_{filler}$ grandma] very kind

Its English translation:

8b) [$_{filler}$ The grandma] [$_{RC}$ who the grandpa helped _$_{gap}$] is very kind.

**Mandarin DCE sentence:**

9a) [$_{N0}$ Jǐngchá] [$_{V1}$ dàibǔ] [$_{RC1}$ _$_{gap1}$ $_{V2}$ ānwèi [$_{RC2}$ _$_{gap2}$ $_{V3}$ mài-le $_{N1}$ fángzi de
[$_{N0}$ The police] [$_{V1}$ arrested] [$_{RC1}$ _$_{gap1}$ $_{V2}$ consoled [$_{RC2}$ _$_{gap2}$ $_{V3}$ sold $_{N1}$ house REL
$_{filler2\ \&\ N2}$ nǎinai] de $_{filler1\ \&\ N3}$ yéyé]
$_{filler2\ \&\ N2}$ grandma] REL $_{filler1\ \&\ N3}$grandpa]
"The police arrested the grandpa who consoled the grandma who sold the house."

**Mandarin incomplete DCE sentence:**

9b) [$_{N0}$ Jǐngchá] [$_{V1}$ dàibǔ] [$_{RC1}$ _$_{gap1}$ $_{V2}$ ānwèi [$_{RC2}$ _$_{gap2}$ $_{V3}$ mài-le $_{N1}$ fángzi de
[$_{N0}$ The police] [$_{V1}$ arrested] [$_{RC1}$ _$_{gap1}$ $_{V2}$ consoled [$_{RC2}$ $_{gap2}$ $_{V3}$ sold $_{N1}$ house REL
$_{filler2\ \&\ N2}$ nǎinai]]
$_{filler2\ \&\ N2}$ grandma]]
"The police arrested (…) consoled the grandma who sold the house."

Using acceptability rating task, Huang and Phillips (2021) found that incomplete DCE sentences, with the head-NP and relativizer omitted, were rated as more acceptable than complete DCE structures, albeit the difference was not statistically significant. This finding implies the presence of a "missing-NP effect" for Mandarin DCE sentences. However, existing theories cannot account for this phenomenon.

*Resource-based account* proposed that the syntactic predictions of a constituent are likely to be "forgotten" if it incurs the highest processing memory load. According to DLT, in a Mandarin DCE sentence the "forgotten" constituent is expected to be the first head-NP "N3" in (9a), which bears the highest integration cost: i) 1 EU for introducing a new discourse referent N3 "grandpa"; ii) 3 EUs for in the integration of N3 "grandpa" and its dependent verb V2 "consoled", with the intervention of 3 discourse referents, "grandma", "house", and "sold" ; iii) 4 EUs for $N_3$ "criminal" and its RC gap, corresponding to the intervention of 4 discourse referents, "grandpa", "grandma", "sold", and "consoled"; iv) and finally 4 EUs for the integration of $N_3$ "criminal" as the object of the sentence and $V_3$ "arrested" in the matrix clause, which is caused by 4 intervening discourse referents, "grandma", "house", "sold", and "consoled". However, this leads to further speculations. As delineated above, the omission of N3 and its relativizer leaves both $RC_1$ verb "V2" and main clause verb "V1" orphaned. While it might be plausible that somehow the processor "forgets" the syntactic prediction from V2, it is highly improbable to "forget" the prediction from main clause verb because of the primacy effect. Adapting *resource-based account* to accommodate the observed missing-NP effect requires the introduction of three additional mechanisms: first, V2 is structurally "forgotten", leading to the "forgetting" of predictions for $N_3$ and relative marker for $RC_1$; second, the predicted sentential object shifts from N3 to N2; third, these two processes must transpire at some point after parsing V2 and before or during parsing N2.

*Interference-based account* faces analogous problems. According to *discrimination hypothesis*, the missing-VP effect arises from competition between similar attachment sites for the integration of the second verb. In the context of Mandarin, Huang and Phillips (2021) proposed that the missing-NP effect transpires when the parser mis-attaches the second post-verb NP to the third post-verb NP position due to the similarity of these two attachment sites. We argue that this analysis violates the recency effect. From an incremental parsing perspective, parsing $V_2$ and $V_3$ in (9a) triggers the predictions for relative clauses, which requires relativizers and head-NPs. When the parser encodes the first *de* and the second post-verb NP, it identifies them as the relativizer and head-NP for relativization with $V_2$ or $V_3$. Given that $V_3$ has been processed more recently, it benefits from a recency effect during relativization.. And normally, parsing the second relativizer and second head-NP prompts the retrieval of $V_2$ for relativization. If the missing-NP effect results from the misattachment of the first head-NP (i.e., the second post-verb NP "$N_2$"), it means that the parser somehow ignores recency effect and retrieves $V_2$ for relativization. This would make $V_3$, a more recently processed verb, orphaned, an outcome that defies logical parsing strategy. Therefore, *interference-based account* should predict no missing-NP effect, since it is unlikely to happen mis-attachment.

*Experience-based account* explains the cross-linguistic difference of the missing-VP illusion based on linguistic statistics. Languages such as German and Dutch, characterized by SOV word order and invariably verb-final relative clauses, enable native speakers to more effectively track verbs in Dutch and German DCE structures compared to English. The occurrence of a missing-VP illusion correlates with the prevalence of verb-final relative clauses. (Futrell et al., 2020). Likewise, the occurrence of a missing-NP illusion should hinge on the frequency of "noun-final" relative clauses. Mandarin relative clause, as demonstrated above, is consistently head (noun) -final. This implicates that parsing Mandarin DCE structure should parallel German and Dutch, where no grammaticality illusion is observed. Huang and Phillips (2021) in Experiment 4 trained Futrell et al.'s (2020) *lossy-context surprisal model* under Mandarin context and found that this model failed to find the missing-NP effect in DCE structure. This suggests that currently *experience-based account* cannot account for this grammaticality illusion in Mandarin.

Finally, let's turn to Huang and Phillips' *interference-and-repair analysis* for a rescue. This analysis, adopting a non-structural perspective, ascribes the missing-NP/VP effect to the possibility of establishing thematic relations among all arguments and predicates within the ungrammatical DCE structure. (see Huang & Phillips, 2021, for details). In the case of missing-NP effect, they suggested that it occurs due to the misattachment of $N_2$ to the $N_3$ position owing to the similarity of the attachment sites. In addition, since in Mandarin DCE sentence sentential object shares the same NP with RC head-NP, they implicitly suggested that this mis-attachment is only to fulfil the thematic relation with main clause verb. This scenario leaves only $V_2$ orphaned. To resolve this issue of orphanage, they introduced a repair mechanism: that is, the parser would somehow notice and "repair" the orphanage of $V_2$, and then attach it to $N_2$, making all predicates connected with at least one argument and therefore resulting in a missing-NP effect.

This *interference-and-repair analysis* appears to adequately describe the Mandarin missing-NP effect. However, this approach presents certain empirical and logical challenges. For starters, the "repair" mechanism seems redundant: if the parser identifies the orphanage of $V_2$, why would it not recognize the incomplete structure as ungrammatical, but instead resort to some additional "repair" process to render it acceptable? Note that both *resource-based account* and *interference-based account* explain missing-VP effect by assuming that the parser somehow unnoticed the orphaned verb. The biggest problem is to speculate that the parser would mis-attach $N_2$ into $N_3$ position only to fulfil thematic-relation with main clause verb, because from a structural perspective, this is a violation of recency effect.[2]

[2] Instead of a mis-attachment analysis, a more plausible one is to assume that due to the primacy effect of the sentential object, $N_2$ is immediately recognized as both the head-NP of $RC_2$ and the object for the main clause. But still, it struggles to explain the orphanage

To sum up, we have argued that no existing theories can sufficiently account for the missing-NP effect in Mandarin DCE sentence. Both *resource-based account* and Huang and Phillips' *interference-and-repair analysis* are marred by excessive speculations. Furthermore, both *interference-based account* and *experience-based account* fall short in predicting a missing-NP effect.

**3 The present study**

Our goal is to address the subsequent research questions:

**RQ1**: What underlies the missing-NP effect?

**RQ2**: Apart from memory constraints, what additional factor could contribute to the processing limitation of Mandarin DCE structure?

For **RQ1**, we propose that the missing-NP effect emerges due to ambiguity in verb interpretation, making incomplete Mandarin DCE structure perfectly grammatical; and that this ambiguity cannot be resolved completely. The rationale is as follows.

First, Mandarin grammar permits the coordination of syntactic objects within the same category without necessitating an overt coordinator. This implies that upon encountering two verbs as in (10), the parser may either embed one within the other, forming an RC structure (embedding interpretation), or conjoin them with a covert coordinator (conjunctive interpretation).

| | | | | |
|---|---|---|---|---|
| (10) | Jǐngchá | dàibǔ | jūliú | xiǎotōu… |
| | The police | arrested | detained | the thief |

Embedding interpretation: The police arrested (someone who) detained the thief.
Conjunctive interpretation: The police arrested and detained the thief.

Second, given that Mandarin RC is head-final, we argue this ambiguity is irresolvable in DCE structure. In a head-initial language such as English, the three NPs in DCE sentence can be separated by overt relative pronouns, precluding any conjunctive interpretation. For Mandarin, however, there is nothing standing in between the verbs to stop them from being coordinated. These two properties conspire to give the incomplete DCE structure (9b) reiterated in (11) with a plausible conjunctive interpretation.[3]

| | | | | | | | |
|---|---|---|---|---|---|---|---|
| (11) | Jǐngchá | [$_{V3}$ dàibǔ] | [$_{RC2\ _gap2}$ $_{V2}$ ānwèi | [$_{RC1\ __gap1}$ | $_{V1}$ mài-le | $_{N1}$ fángzi | de |
| | The police | [$_{V3}$ arrested] | [$_{RC2\ _gap2}$ $_{V2}$ consoled | [$_{RC1\ __gap1}$ | $_{V1}$ sold | $_{N1}$ house | REL |
| | $_{filler1\ \&\ N2}$ nǎinai] | | | | | | |
| | $_{filler1\ \&\ N2}$ grandma] | | | | | | |

conjunctive interpretation: ✔The police arrested (and) consoled the grandma who sold the house.
embedding interpretation: *The police [arrested [--- consoled]] the grandma who sold the house.

Clearly the conjunctive interpretation has awkward semantics. But given the fact that DCE structure is highly complex, the parser might somehow ignore this awkwardness, and from the perspective of computational efficiency, turn to a

---

of V2.

[3] There are ways to cue parser the existence of a relative clause (e.g., Jäger et al., 2015; Lin and Bever, 2011). We argue in Section 6 that these ways to deal with ambiguity in Mandarin single RC may not be viable for DCE structure.

conjunctive interpretation, since building an embedded interpretation is more resource-intensive, as expounded below.

If the above assumption is on the right track, then the missing-NP effect reported by Huang and Phillips (2021) cannot be regarded as a "grammaticality illusion". The only difference between the incomplete and complete DCE sentences lies in the interpretation of the verbs. Furthermore, we can answer **RQ2** by venturing that the processing limitation of Mandarin DCE structure is primarily driven by the ambiguity in interpreting the verbs. That is, at parsing sentences (i.e., DCE sentence) of high complexity, given the choice of embedding one syntactic object into the other, or conjoining the two, the parser would initially take the second route to build a representation. As illustrated in (12), if no strings attached, building a representation for an unfinished sequence "NVVVN…" takes fewer steps via conjunctive interpretation than via embedding interpretation; the "steps" are quantified by times of MERGE.

This speculation aligns with the garden-path theory (Frazier, 1987; Frazier & Fodor, 1978) and its related *Minimal Commitment* models (Frazier & Rayner, 1982; Just & Carpenter, 1980; Mitchell, 1994). Before parsing a relativizer *de*, the parsing process is presumably governed by a *Minimal Attachment Strategy* (Frazier & Fodor, 1978), where each received lexical item is attached into the existing structure with fewest possible nodes. The parser conjoins the verbs at some point before encountering a relativizer and maintains this conjunctive interpretation until the emergence of a relativizer "*de*" that bars such interpretation, thereby engendering a garden path effect. This may account for the observation that Mandarin native speakers assigned higher acceptability ratings to the missing-NP DCE structure compared to the complete DCE structure in Huang and Phillips' acceptability rating task. Both structures being syntactically correct, the complete DCE structure takes more steps, therefore higher memory demands, to build a representation than does the missing-NP structure. This led participants to perceive the complete DCE structure as more cognitively taxing, resulting in relatively lower ratings.

(12) a. conjunctive interpretation = 6 MERGEs

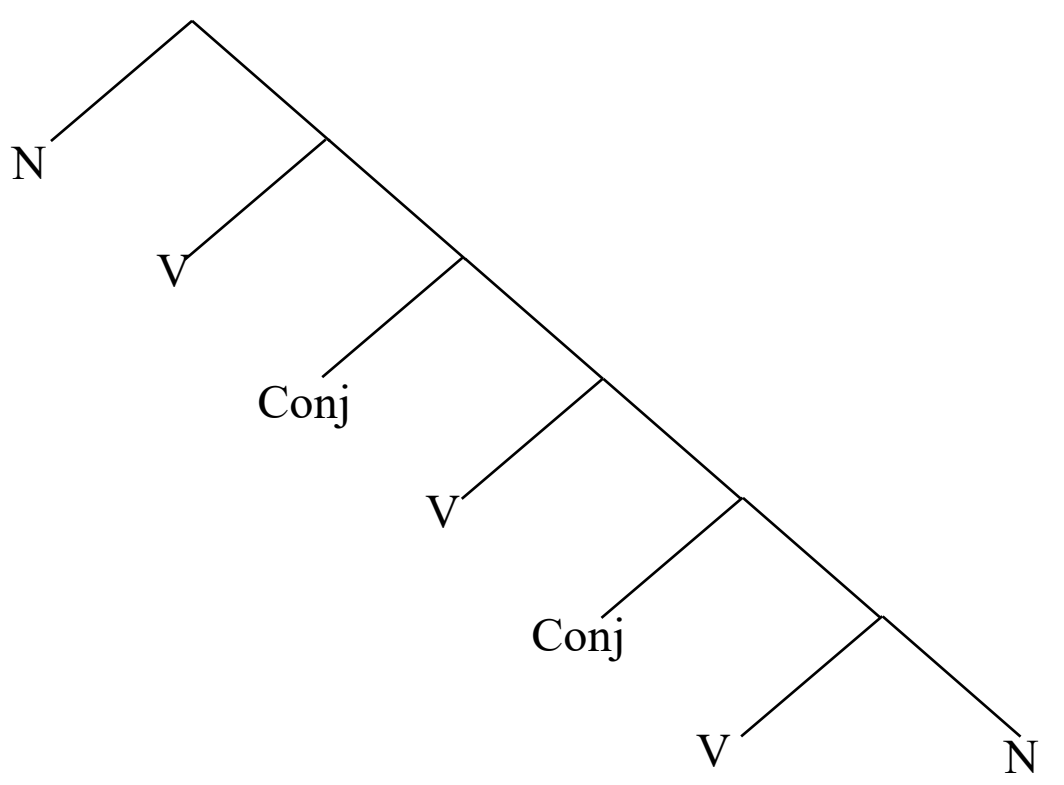

b. embedding interpretation = 8 MERGEs

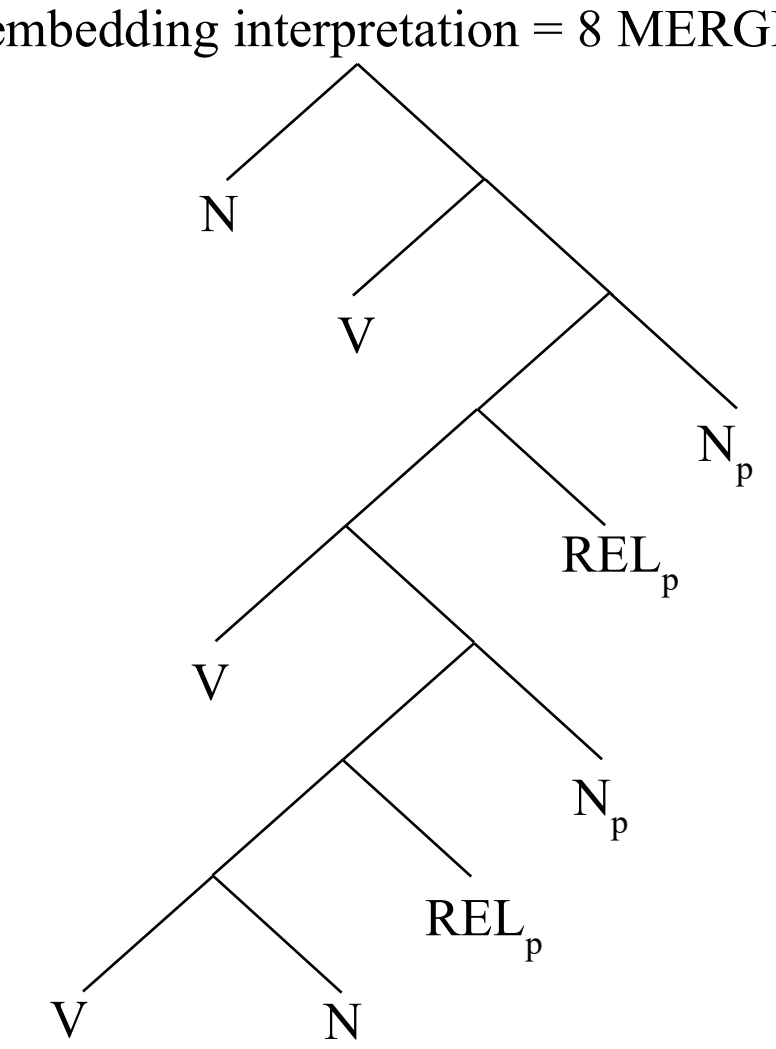

How can we substantiate the above assumptions? One plausible solution is to assess whether a less complex structure endures this ambiguity. The rationale is that if processing a less complex structure bears ambiguity, then a similar but more complex structure should undoubtedly encounter the same ambiguity. Consequently, if ambiguity in interpreting verbs is present in a structure like (13), which is less complex than Mandarin DCE structure due to the position of its self-embedding RCs then parsing Mandarin DCE structure should also exhibit this ambiguity.

(13) a. **Complete quasi-DCE sentence**

[$_{\text{matrix clause}}$[$_{\text{RC2 _gap2}}$ $_{\text{V2}}$ dǎ-le [$_{\text{RC1 __gap1}}$ $_{\text{V1}}$ bāngzhù $_{\text{N1}}$ nǎinai de $_{\text{filler1 \& N2}}$ yéyé]
[$_{\text{matrix clause}}$[$_{\text{RC2 _gap2}}$ $_{\text{V2}}$ hit [$_{\text{RC1 __gap1}}$ $_{\text{V1}}$ helped $_{\text{N1}}$ grandma REL $_{\text{filler1 \& N2}}$ grandpa]

de $_{\text{filler2 \& N3}}$ zuìfàn] $_{\text{matrix verb}}$ hē-le shuǐ]
REL $_{\text{filler2 \& N3}}$ the criminal] $_{\text{matrix verb}}$ drank water]

"The criminal that hit the grandpa that helped the grandma drank water."

b. **Incomplete quasi-DCE sentence**

[$_{\text{matrix clause}}$[$_{\text{RC2 _gap2}}$ $_{\text{V2}}$ dǎ-le [$_{\text{RC1 __gap1}}$ $_{\text{V1}}$ bāngzhù $_{\text{N1}}$ nǎinai de $_{\text{filler1 \& N2}}$ yéyé]
[$_{\text{matrix clause}}$[$_{\text{RC2 _gap2}}$ $_{\text{V2}}$ hit [$_{\text{RC1 __gap1}}$ $_{\text{V1}}$ helped $_{\text{N1}}$ grandma REL $_{\text{filler1 \& N2}}$ grandpa]
$_{\text{matrix verb}}$ hē-le shuǐ]

$_{\text{matrix verb}}$ drank water]

Conjunctive interpretation: ✔The grandpa that hit and helped the grandma drank water.
Embedding interpretation: * --- that hit the grandpa that helped the grandma drank water.

If our line of reasoning is on the right track, native Mandarin speakers influenced by this ambiguity should perceive the incomplete sentence (13b) as grammatical, hence demonstrating a "missing-NP effect". In addition, parsing (13a) should experience a garden-path effect at the second relativizer, which adds processing load. Behavioural measures such as acceptability/comprehensibility rating tasks may not yield this information, as low ratings for incomplete structures could stem from the awkward semantic combination of the verbs. Most studies concerning the processing limitation of DCE structure and its grammaticality illusion employed these behavioural measurements (e.g., Frank & Ernst, 2019; Gibson & Thomas, 1999; Gimenes et al., 2009; Huang & Phillips, 2021). A more effective method is to measure participants' brain responses.

We recorded and analysed event-related potentials from native Mandarin speakers reading both complete and incomplete quasi-DCE sentences. Should the incomplete sentence be perceived as ungrammatical, a centro-posterior P600 effect, typically indicative of syntactic anomalies (Friederici et al., 2002; Hagoort et al., 1993; Neville et al., 1991; Osterhout & Holcomb, 1992), should be observed when the parser encounters the main clause verb. Likewise, if an incomplete quasi-DCE sentence is seen as grammatical, then no sign of P600 effect that reflects syntactic anomaly would be expected at the main clause verb, indicating the presence of missing-NP effect. Additionally, as some studies have posited, the ease of integrating words into the preceding conceptual context may generate N400 with varying amplitudes (Friederici, 2002, 2017; Hagoort, 2005). We anticipate that in this case, a more pronounced N400 may be observed at the main clause verb in incomplete sentences, attributable to the awkward semantics of the conjunctive interpretation. Also, if ambiguity underlies the processing limitation of (quasi) DCE structures, a garden-path effect should be anticipated upon parsing the second relativizer (compared to the first one). This could possibly be reflected by a *left-anterior negativity* (LAN) indicating an increase of working memory (Fiebach et al., 2001; King & Kutas, 1995; Vos et al., 2001), or a more frontally distributed P600 effect suggesting syntactic reanalysis (Friederici et al., 2002; Kaan & Swaab, 2003).

## 4 Experiment 1

### Materials

80 pairs of target sentences were constructed, each encompassing both "complete" and an "incomplete" quasi-DCE sentences, and segregated into two test lists. No target sentence appeared more than once within a single list. Sentences in the "incomplete" condition were derived from the "complete" condition by substituting the second relative marker "*de*" with the coordinator "*he*", exemplified in (14b). This ensured that sentences in both conditions had an equal number of characters and words in the self-embedding RC region (exactly 12 characters, 7 words) so as to mitigate any potential influence ERP responses attributable to clause length (Van Petten & Kutas, 1991). The average frequency of words is 1.84 fpm (frequency per million words) , based on the corpora from BLCU Corpus Centre (BCC). We particularly focus on the comparison of main clause verbs in complete and incomplete sentence, of relativizers in complete sentence and of head-NPs in complete sentence. The average frequency of the first head-NP is 0.45 fpm and that of the second head-NP is 0.56 fpm, with no significant difference between them after a two-sample *t*-test ($t(158) = -.926$, $p = .356$).

(14) a. **Complete quasi-DCE sentence**

| | | | | | |
|---|---|---|---|---|---|
| [$_{\text{matrix clause}}$[$_{\text{RC2 _gap2}}$ | $_{\text{V2}}$ dǎ-le | [$_{\text{RC1 __gap1}}$ $_{\text{V1}}$ bāngzhù | $_{\text{N1}}$ xiǎomāo | de | $_{\text{filler1 \& N2}}$ xiǎogǒu] |
| [$_{\text{matrix clause}}$[$_{\text{RC2 _gap2}}$ | $_{\text{V2}}$ hit | [$_{\text{RC1 __gap1}}$ $_{\text{V1}}$ helped | $_{\text{N1}}$ the cat | REL | $_{\text{filler1 \& N2}}$ the dog] |

| | | | |
|---|---|---|---|
| de | $_{\text{filler2 \& N3}}$ xiǎohóu] | $_{\text{matrix verb}}$ hē-le | $_{\text{OBJ}}$ shuǐ] |
| REL | $_{\text{filler2 \& N3}}$ the monkey] | $_{\text{matrix verb}}$ drank | $_{\text{OBJ}}$ water] |

"The monkey that hit the dog that helped the cat drank water."

b. **Incomplete quasi-DCE sentence**

| | | | | | |
|---|---|---|---|---|---|
| [$_{\text{matrix clause}}$[$_{\text{RC2 _gap2}}$ | $_{\text{V2}}$ dǎ-le | [$_{\text{RC1 __gap1}}$ $_{\text{V1}}$ bāngzhù | $_{\text{N1}}$ xiǎomāo | de | $_{\text{filler1 \& N2}}$ xiǎogǒu |
| [$_{\text{matrix clause}}$[$_{\text{RC2 _gap2}}$ | $_{\text{V2}}$ hit | [$_{\text{RC1 __gap1}}$ $_{\text{V1}}$ helped | $_{\text{N1}}$ the cat | REL | $_{\text{filler1 \& N2}}$ the dog |

| | | | | |
|---|---|---|---|---|
| he | $_{\text{filler2 \& N3}}$ xiǎohóu] | ---] | $_{\text{matrix verb}}$ hē-le | $_{\text{OBJ}}$ shuǐ] |
| and | $_{\text{filler2 \& N3}}$ the monkey] | ---] | $_{\text{matrix verb}}$ drank | $_{\text{OBJ}}$ water] |

Conjunctive interpretation: The monkey and the dog that hit and helped the cat drank water.
Embedding interpretation: * --- that hit the dog and the monkey that helped the cat drank water.

In Mandarin *de* functions not only as a relativizer, but also as a possessive marker, meaning that RC$_1$ in (13a), reiterated in (15), might be construed possessively:

(15)

| | | | |
|---|---|---|---|
| bāngzhù | nǎinai | de | yéyé |
| helped | grandma | DE | grandpa |

Possessive interpretation: *pro* helped grandma's grandpa.
Relativization interpretation: the grandpa who helped the grandma

To reduce the possibility of a possessive interpretation at RC$_1$, we chose non-human animals (e.g. dog, cat, etc.) as NPs, as they seem less likely to have a possessive interpretation with *de* connecting them.[4]

(16)

| | | | |
|---|---|---|---|
| bāngzhù | xiǎomāo | de | xiǎogǒu |
| helped | the cat | DE | the dog |

Possessive interpretation: ? *pro* helped the cat's dog.
Relativization interpretation: the dog who helped the cat

---

[4] Another way of resolving this local ambiguity in a single RC structure is to add adverbs between first post-verb noun and *de* (Jäger et al., 2015):

(i)

| | | | | |
|---|---|---|---|---|
| bāngzhù | nǎinai | jǐcì | de | yéyé |
| helped | grandma | several times | DE | grandpa |

"the grandpa who helped the grandma several times"

This, however, would induce additional integration cost when there is self-embeddings of RCs, as illustrated in (ii), where the adverb *jǐcì* "several times" forms a long nested-dependency with RC$_2$ verb:

(ii)

| | | | | |
|---|---|---|---|---|
| [$_{\text{RC2 _gap2}}$ dǎ-le | [$_{\text{RC1}}$ …] | jǐcì | de | yéyé] |
| [$_{\text{RC2 _gap2}}$ hit | [$_{\text{RC1}}$ …] | several times | DE | grandpa] |

"the grandpa who hit the grandma several times [$_{\text{RC1}}$ …]"

Clearly, this extra integration cost will increase additional processing limitation for our target structure. Therefore, we did not adopt this disambiguation solution.

In alignment with Huang and Phillips (2021), the verbs within the RC regions were meticulously chosen to favour NP complements, thereby preventing the occurrence of both verbs exclusively accepting clause-like complements, such as *xǐhuān* (like), and that they have distinct lexical semantics and grammatical aspect, with only the first verb having a past tense aspect *le*/*guo*.

In addition, 80 filler sentences of comparable length and complexity were formulated. They all contained relative clauses to amplify the likelihood of interpreting Mandarin "*de*" as a relative marker (Lin & Bever, 2011). Grammatical fillers constituted 50% of the total, based on the assumption that participants would regard sentences in the incomplete condition as ungrammatical. All grammatical fillers have a coordinator *he* in between two NPs, ensuring that participants' judgments on grammaticality were not solely on the presence of a coordinator.

**Participants**

Twenty-four native speakers of Mandarin Chinese were recruited from a university located in south-eastern China. Written informed consents were signed by all participants before the experiment. All participants rewarded with extra credits after they completed the experiment (19 females, 5 males; age 18-25 ys, mean = 19.61, SD = 1.83). They all have normal or corrected-to-normal vision and did not report any neurological disorders or heart diseases. The experimental protocol was approved by the Ethics Committee of Soochow University, China.

**Procedure**

Participants were seated in a comfortable chair in a sound-attenuated room, 60 cm away from the monitor. Psychtoolbox-3 was installed on MATLAB to present stimuli with custom-written scripts (Kleiner et al., 2007). Prior to the experiment, participants were told to read carefully the instructions and went through a practice session to familiarize themselves with the procedure and, if needed, repeat this session.

The target sentences were presented with the Rapid Serial Visual Presentation (RSVP) paradigm and the passive reading paradigm. This means that we did not include any behavioural tasks for reasons that tasks like acceptability/grammaticality rating may affect ERP responses (Schacht et al., 2014), and that additional tasks may impose additional cognitive load on participants , considering the inherent complexity of the reading materials.

Each trial began with a 20-point-font fixation cross on the centre of the monitor and would not disappear until the participant pressed the SPACE bar. This aimed to provide participants with a brief respite, mitigating fatigue throughout the experiment. Then, each 50-point-font Chinese simplified word with white colour was presented on the centre of a black background with 300ms duration and 500ms ISI, making an 800ms stimuli onset asynchrony (SOA). This time interval is based upon previous studies that focused on Mandarin SRC/ORC processing difference (Bulut et al., 2018; Packard et al., 2011; Qiu & Zhou, 2012; Xiaoxia et al., 2016; Xiong et al., 2019; Xu et al., 2024; Yang & Perfetti, 2006; Yang et al., 2010). The length of words was controlled to exceed no more than 3 Chinese characters to reduce the change of visual angle across displays (as shown in Fig.). The final word was followed by a 1000ms blank screen and then enters the next trial.

Participants were instructed to minimize blinks, eye movements, and muscle movements during reading to diminish artifact occurrence. In total, they engaged in four blocks of sentence reading, each encompassing 40 sentences. All sentences were first randomized and then randomly dispatched to a block. The aggregate duration of the experiment, including preparatory activities, approximated two hours.

**EEG recording and preprocessing**

The continuous scalp EEG signals were recorded from 64-channel wireless EEG system (NeuSen. W 64, Neuracle, Changzhou, China), placed at the standard extended 10-20 system. The EEG data were online-referenced to CPz and grounded at FPz with a sampling rate at 1000 Hz. Bipolar electrodes were placed at left and right mastoids for later offline re-reference. The impedances were kept under 20 kΩ for all electrodes and less than 5 kΩ for the mastoid sites.

The continuous EEG data were preprocessed by EEGLAB (Delorme & Makeig, 2004) and ERPLAB (Lopez-Calderon & Luck, 2014). The 50 Hz line noise was identified by visual inspection of channel spectra and removed using Parks-McClellan Notch filter. At the first stage of preprocessing, the data were filtered offline (IIR Butterworth non-casual filters) with a bandpass filter set at 0.1-30 Hz (roll-off 24 dB/octave, order 4). Then an independent-component analysis (ICA) was performed on the continuous data to correct ocular and muscle artifacts (Delorme et al., 2005).

After that, the data were re-referenced offline to the average of left and right mastoids and segmented into epochs which are time-locked to the onset of the critical words according to pre-assigned bins. Each epoch had a 1000ms duration, with 200ms pre-stimulus baseline, 300ms word duration and 500ms ISI. Then we performed artifact detection in epoched data using functions in ERPLAB, with moving window peak-to-peak threshold set at 100 µV (moving windows full width at 200ms, window step at 20ms). Meanwhile, manual channel scroll was also applied for other artifacts detection. After that, epochs that were highlighted by artifact detection were excluded from further analysis. The raw data of three participants were excluded due to excessive artifacts.

**Statistical analysis**

To address the issue of implicit multiple comparison (for detailed description of this issue, see Luck, 2014, Chapter 10; Luck & Gaspelin, 2017), a non-parametric cluster-based permutation test was conducted (Bullmore et al., 1999; Maris & Oostenveld, 2007). This massive univariate analysis was used to replace the conventional mean amplitude analysis for several reasons. First, traditional ERP analysis methods, like visual inspection of ERP waves, have faced criticism for potentially exacerbating the multiple comparison problem through subjective and implicit selection of time windows and electrodes based on observation of the waveforms, while those significant effects drawn by such procedure might be "bogus", thereby increasing Type I error (Fields & Kuperberg, 2020; Luck, 2014; Luck & Gaspelin, 2017). Second, selecting specific time windows and electrodes based on a priori measurement parameters (i.e. previous studies) may increase the rate of Type II error especially when looking for ERP effects that are small and long-lasting, since a same effect may have a different latency owing to different experimental design or environment. Third, compared with the conventional mean amplitude ANOVA method, massive univariate analysis provides better temporal and spatial resolution while maintaining a reasonable correction for the multiple comparison. Meanwhile, the cluster-based permutation test has been shown to have great power for discovering broadly distributed ERP effects like the P600 than other massive univariate analyses (Fields & Kuperberg, 2020; Groppe et al., 2011; Maris & Oostenveld, 2007).

The non-parametric cluster-based permutation test was performed using Mass Univariate ERP Toolbox (Groppe et al., 2011). We first downsampled the data from 1000Hz to 100Hz using the *boxcar* filter to reduce the overall number of comparisons (Luck, 2014). Then ANOVAs were performed on each electrode site and time point in the selected time window. The *t*-values greater than a threshold of these spatially and temporally distinct samples were added for cluster inclusion. We set this threshold *p*-value at 0.05. Electrodes within approximately 5.44cm from one another were considered as spatial neighbors (mean = 6.5, *SD* = 1.5) and adjacent time points as temporal neighbors. After that, all spatially adjacent electrodes with the similar significant time points were grouped into a single cluster. Finally, clusters' *t*-values were summed and examined by permutation-based corrections for multiple comparisons, with alpha level at 0.05 and a recommended 100,000 number of permutations (Fields, 2017).

To maximize statistical power, we first set a priori time window of 500-800 ms to detect any P600 effect across these three pairs of comparisons, as suggested by Groppe et al. (2011). This time window was grounded on prior studies reporting a P600 effect in parsing relative clauses (Bulut et al., 2018; Friederici et al., 2002; Hagoort et al., 1993; Kaan et al., 2000; Xiong et al., 2019). Brain potentials at each electrode will first be averaged within this time window and then analysed based on cluster-based permutation test. Subsequently, we performed an exploratory analysis that tests all time point of interest within a time window of 0-800ms to identify any unexpected effect.

**Results**

The artifact removal process excluded 9.86% of trials on average, from participants included for further analysis. There was no significant difference in rejection rates among the conditions (Tukey correction, $F(5, 120) = .143$, $p = .982$).

*Priori mean time window (500-800 ms) looking for P600 effect*

At the main clause verbs, the mean time window cluster-based permutation test reported two positive clusters, yet both were non-significant (*ps* > .15). This finding suggests the absence of a P600 effect for the incomplete quasi-DCE sentence within this time window.

For the comparison of relativizers *de* in the complete quasi-DCE structure, the test found a significant positive cluster (cluster-mass t = 98.1323, $p < .01$, $\alpha = .05$). This cluster exhibited a centro-posterior distribution with its spatial mass peak localized at C4. And at the RC head-NPs in the complete quasi-DCE structure, the test also found a significant positive cluster (cluster-mass t = 96.7799, $p = .012$, $\alpha = .05$) that has a broad scalp distribution. Its spatial mass peak is found at electrode TP7. Since Mandarin RC head-NP is posited after the relativizer *de*, the presence of a relativizer would instigate the anticipation of a head-NP (Packard et al., 2011; Xiong et al., 2019). Hence, the P600 effect at the head-NPs is likely to indicate an increase of structural integration cost, with the second head-NP requiring more resources to fulfil its filler-gap dependency than the first head-NP.

*Exploratory analysis (0-800 ms)*

To explore any other potential effects beyond the predefined time window, we conducted the test on all time points.

At the main clause verbs, the exploratory test identified one significant negative cluster (cluster-mass t = -891.8865, $p = .045$, $\alpha = .05$). The cluster has a centro-posterior distribution and a 160-400ms latency, with its spatial-temporal mass peak over electrode TP8 around 290ms. Fig. 1 displays the grand-averaged wave form of TP8, the topography of the difference potentials (incomplete – complete) and the raster plot of electrodes included in this cluster. We interpret this effect as N400 as the latency and the scalp distribution of it matches that of N400 (Swaab et al., 2011).

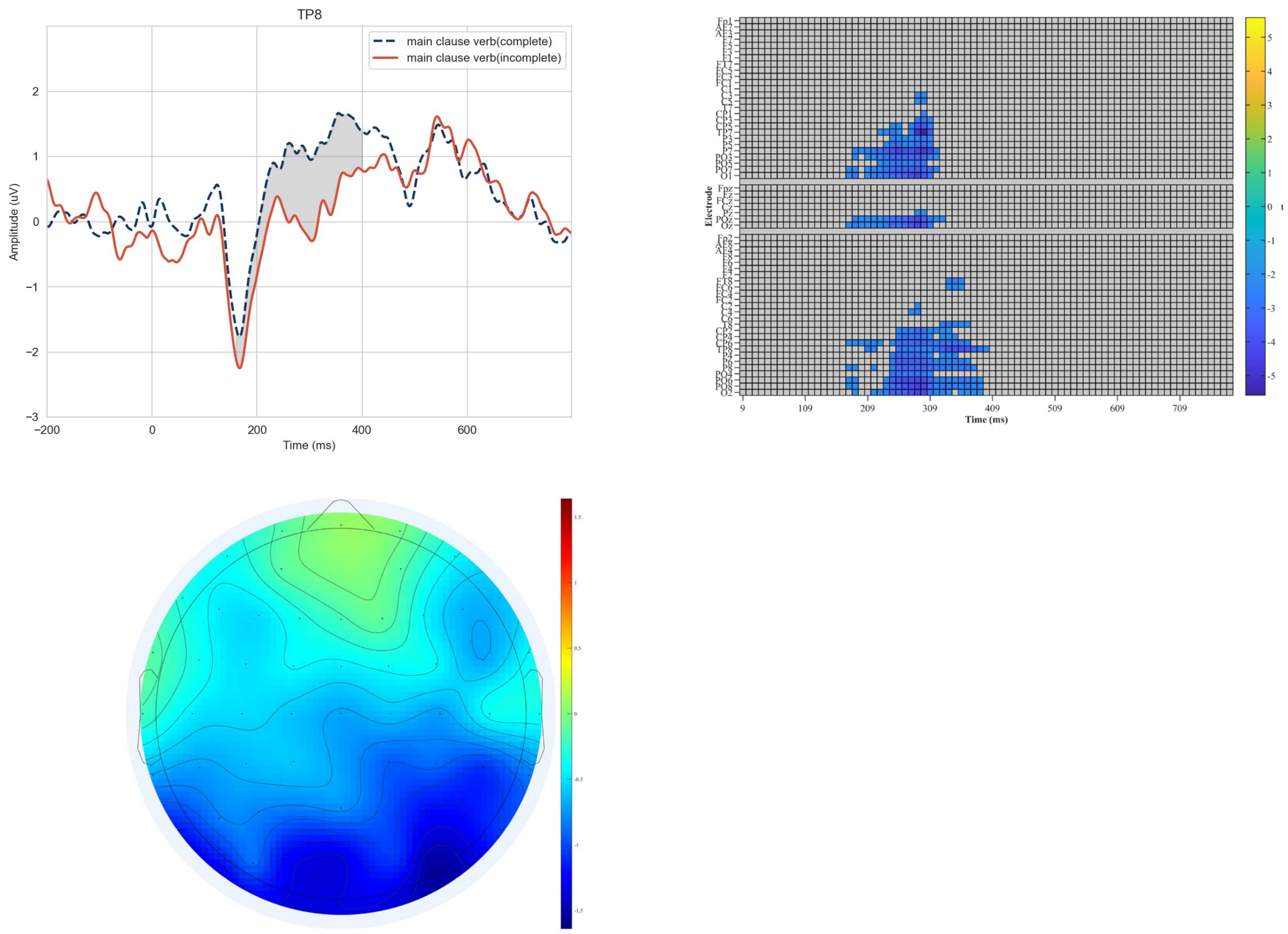

Fig. 1 Cluster mass exploratory results from main clause verbs comparison. This graphic includes: a) the grand-averaged waveforms of electrode TP8, where the effect is maximal; b) the raster plot showing the spatial-temporal distribution of this effect; c) the topography of the difference wave (incomplete – complete) with averaged amplitude within a 190-400 ms time-window.

For the first and second relativizers comparison, our exploratory test found a significant positive cluster (cluster-mass t = 3059.9022, $p < .01$, $\alpha = .05$). This effect has a long-lasting latency of 180-800 ms, which includes the 500-800ms priori time window, with its spatial-temporal mass peak over electrode CP1 around 640ms. Electrodes in posterior region are more consistently involved than in anterior region. Fig. 2 shows grand-averaged waveform of CP1, the raster plot and the topographies of the difference wave (second – first relativizers) in a stepwise time interval of 200 ms.

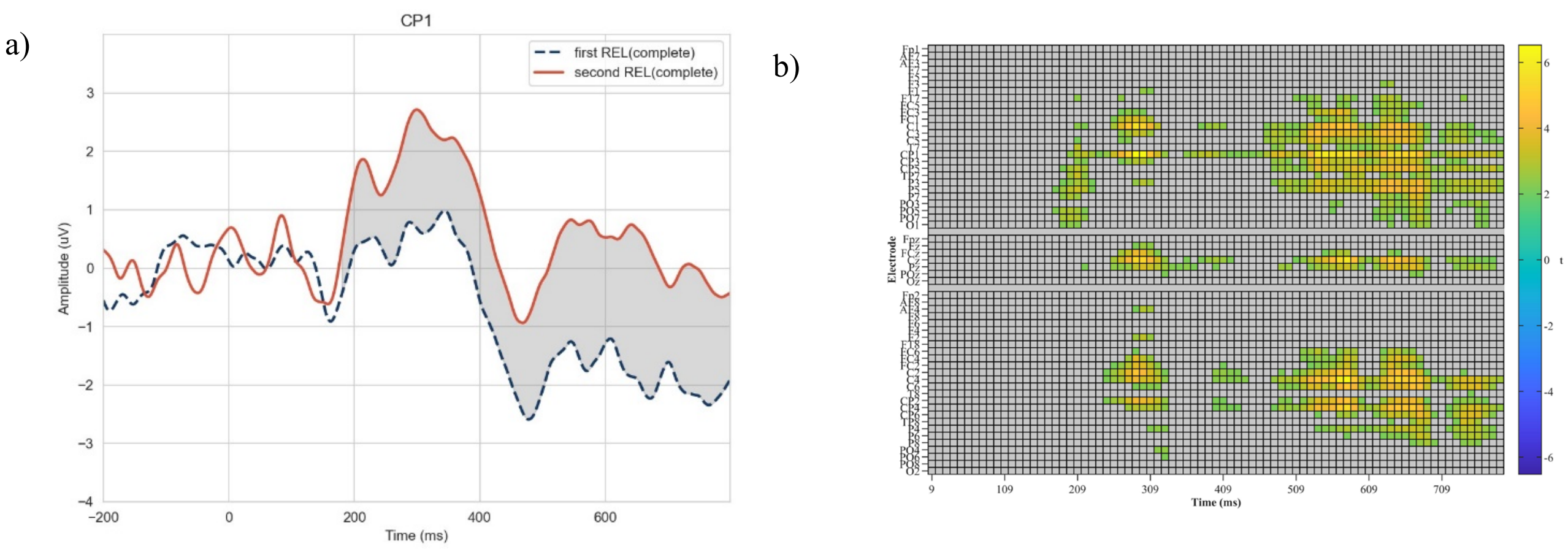

c)

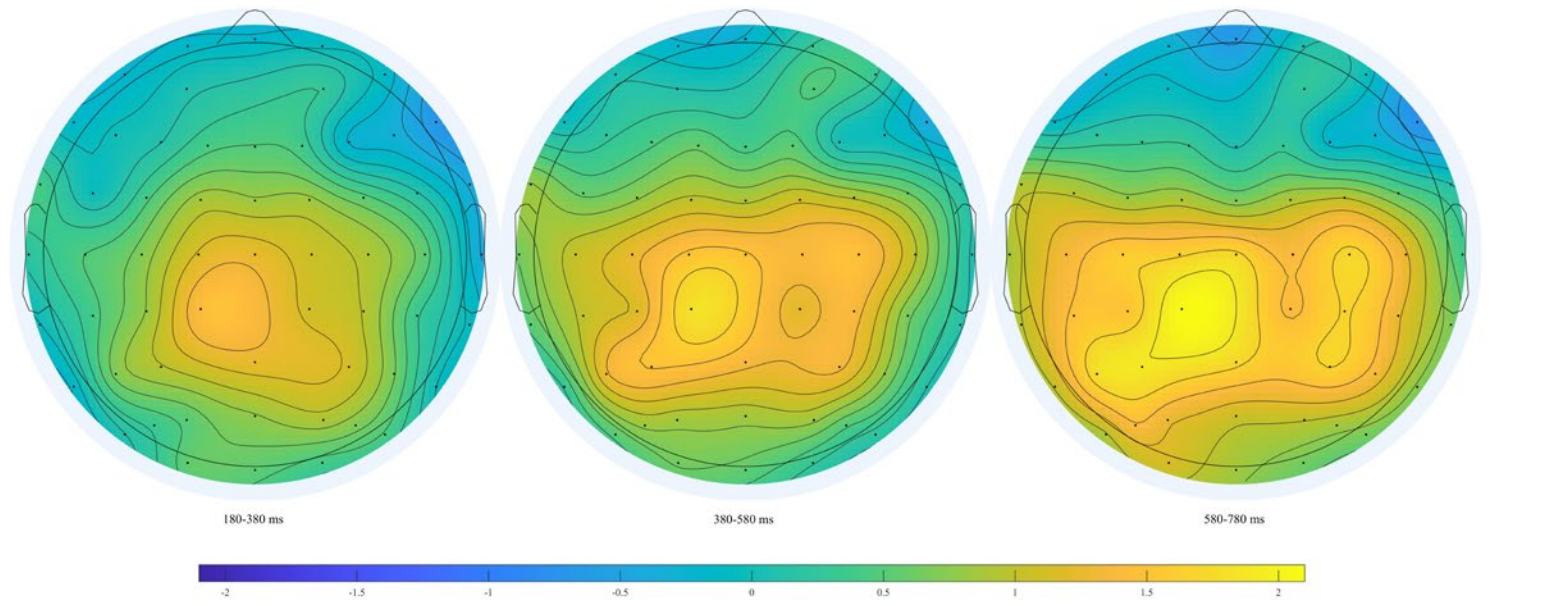

Fig. 2 Cluster mass exploratory results from relativizers comparison. This graphic contains: a) grand-averaged waveform of electrode CP1 where the effect is the strongest; b) the raster plot showing the spatial-temporal distribution of the effect; and c) topographies of mean amplitude of difference ERPs (second – first relativizers) with a stepwise time interval of 200ms.

For the comparison between first and second RC head-NPs, the test found two significant positive cluster (cluster-mass t = 2724.7221, $p < .01$, $\alpha = .05$). The first cluster is featured by a latency of 380-610ms and a broad distribution, with its spatial-temporal mass peak over electrode FC4 around 510ms. The second cluster also has a broad distribution with its spatial-temporal mass peak over TP7 around 770ms. Its latency is 620-800ms. Fig. 3 shows the grand-averaged waveform of FC4 and TP7, the raster plot, and the topographies of the difference wave (second – first head-NPs) in a stepwise time interval of 100 ms.

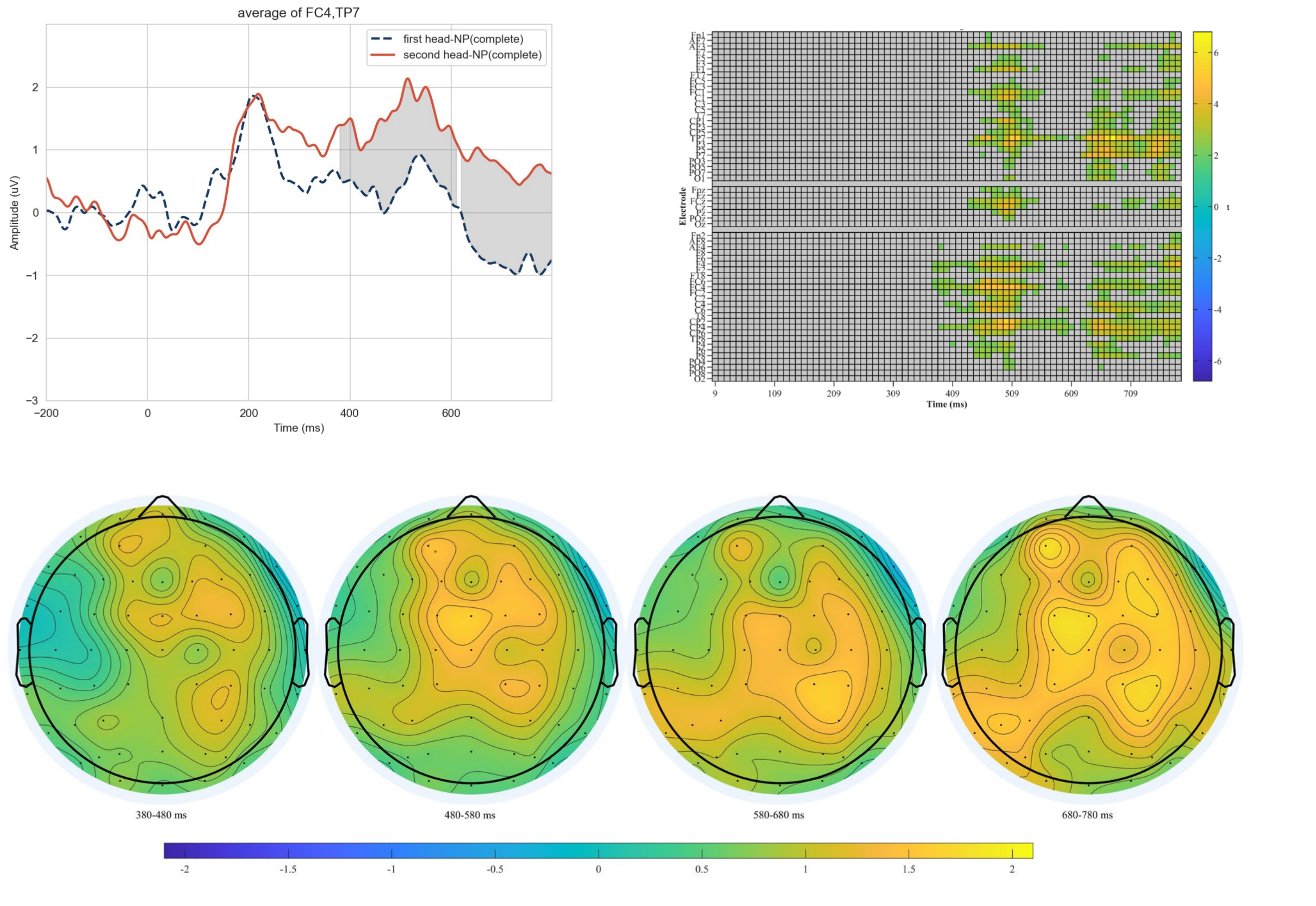

Fig. 3 Cluster mass exploratory results from head-NPs comparison. This graphic includes: a) grand-averaged waveform of electrode FC4 and TP7 where the effect is maximal; b) raster plot showing the spatial-temporal distribution of the effect; and c) topographies of mean amplitude of difference potentials (second – first head-NPs) with a stepwise time interval of 200ms.

**Discussion**

Experiment 1 confirmed that "missing-NP effect" exists in Mandarin quasi-DCE structure, whose complexity is maneuvered to be lower than that of DCE structure by placing the self-embedding RC region into the subject position of the main clause. This is evidenced by the non-significant difference in the priori time window looking for P600 effect between the two conditions at the main clause verbs. In addition, the N400 effect found by the exploratory analysis suggests that the parser perceives the incomplete quasi-DCE sentence as semantically incongruent.

The results pertaining to the main clause verbs can be interpreted within the framework of our *ambiguity-based analysis*. Somewhere between the first encountered verb and first head-NP, the parser conjoined the received two verbs and maintained this conjunctive interpretation of the verbs. If no additional relativizer is encountered, which is the case for our incomplete condition, the parser ends the interpretation of the self-embedding RC region and start processing the main clause. Given that no syntactic violation is encountered during the process of the RC region, no P600 effect that reflects a syntactic-repair process is observed; furthermore, due to the distinct lexical semantics and grammatical aspects of the verbs in our target sentences, the parser noticed the semantic incongruency of the conjunctive interpretation, thereby eliciting a N400.

Likewise, if a second relativizer is received, which is the case for complete condition, a garden-path effect will be experienced since the conjunctive interpretation fails to accommodate a second relativization process. Consequently, at this point, the parser structurally reanalyses the representation, aligning with the embedding interpretation of the verbs. This is evidenced by the centro-posterior P600 effect at the second relativizer, compared to the first. Although reanalysis-related P600, indicative of garden-path effects, is purported to be more frontally distributed (Friederici et al., 2002; Kaan & Swaab, 2003), there is evidence suggesting that a posteriorly distributed P600 may also signal structural reanalysis. For instance, Xiong et al. (2019) found that the relativizer in Mandarin object-extracted RC elicited a centro-posterior distributed P600, compared to that in subject-extracted RC. They interpreted this effect as reflecting structural reanalysis, which renders Mandarin ORC more difficult to process than SRC.

Nevertheless, *interference-and-repair based analysis* can offer alternative explanation for these findings. First, as to the non-significance of P600 at the main clause verbs, it could assume that the syntactically "orphaned" verb in the incomplete quasi-DCE structure can be thematically combined with the first head-NP by a repair mechanism, which makes all the predicates linked to at least one argument. Mandarin native speakers, therefore, exploiting this shallow representation may perceive the structurally incomplete quasi-DCE structure as "thematically" complete. Second, the P600 effect found at the second relativizer may simply indicate an increase of structural integration cost (Kaan et al., 2000)—integrating the second relativizer into the existing structure consumes comparatively more resources than the first. This poses challenge to our structural-reanalysis interpretation of the P600 effect.

Apparently, Experiment 1 cannot give conclusive answer to what induces the missing-NP effect in Mandarin (quasi-) DCE structure. To rule out *interference-and-repair based analysis*, we must determine: Q1) Do native Mandarin speakers experience a missing-NP effect when parsing (quasi-) DCE sentences with unambiguous verb region? Q2) will a similar P600 effect be observed at the second relativizer in such sentence?

As we have discussed, due to Mandarin RC's head-final property, ambiguity at verbs in Mandarin DCE structure cannot be resolved until the presence of relativizers. However, there might be some ways to reduce this ambiguity. For instance, altering the lexical semantics of verbs to ensure each verb associates with an NP-complement of distinct animacy status could be a solution. As illustrated in (12a), conjunctive interpretation of verbs requires the received verbs sharing the same NP complement. If we design quasi-DCE sentences in a way that the two RC verbs select their NP complements with different animacy status, the parser may notice this *semantic cue* and choose to give an embedding interpretation for the two verbs. For instance, the verb “assault” can only select an animate NP (e.g. a person) but not an inanimate one (e.g. a house). It is semantically/pragmatically implausible to conjoin the verb “assault” with a verb (e.g. demolish) that only s-selects an inanimate NP complement. If the above assumption is on the right track, according to our *ambiguity-based analysis* we should expect sentences in (17) with a reduced ambiguity at the RC verb region such that the parser would initially give an embedding interpretation for the RC verbs, thereby creating little or no missing-NP effect at the end of the RC region in (17b) and much less processing difficulty for (17a) because of the absence of a garden-path effect.

(17) a. **Complete quasi-DCE sentence with reduced ambiguity**

[$_{\text{matrix clause}}$[$_{\text{RC2}}$ _$_{\text{gap2}}$ $_{\text{V2}}$ ōudǎ [$_{\text{RC1}}$ __$_{\text{gap1}}$ $_{\text{V1}}$ mài-le $_{\text{N1}}$ fángzi de $_{\text{filler1 \& N2}}$ nǎinai]
[$_{\text{matrix clause}}$[$_{\text{RC2}}$ _$_{\text{gap2}}$ $_{\text{V2}}$ hit [$_{\text{RC1}}$ __$_{\text{gap1}}$ $_{\text{V1}}$ sold $_{\text{N1}}$ the house REL $_{\text{filler1 \& N2}}$ the grandma]

de $_{\text{filler2 \& N3}}$ yéyé] $_{\text{main clause verb}}$ hē-le shuǐ]
REL $_{\text{filler2 \& N3}}$ the grandpa] $_{\text{main clause verb}}$ drank water]

“The grandpa that hit the grandma that sold the house drank water.”

b. **Incomplete quasi-DCE sentence with reduced ambiguity**

[$_{\text{matrix clause}}$[$_{\text{RC2}}$ _$_{\text{gap2}}$ $_{\text{V2}}$ ōudǎ [$_{\text{RC1}}$ __$_{\text{gap1}}$ $_{\text{V1}}$ mài-le $_{\text{N1}}$ fángzi de $_{\text{filler1 \& N2}}$ nǎinai]
[$_{\text{matrix clause}}$[$_{\text{RC2}}$ _$_{\text{gap2}}$ $_{\text{V2}}$ hit [$_{\text{RC1}}$ __$_{\text{gap1}}$ $_{\text{V1}}$ sold $_{\text{N1}}$ the house REL $_{\text{filler1 \& N2}}$ the grandma]
he $_{\text{filler2 \& N3}}$ yéyé] ---] $_{\text{main clause verb}}$ hē-le shuǐ]
and $_{\text{filler2 \& N3}}$ the grandpa] ---] $_{\text{main clause verb}}$ drank water]

Conjunctive interpretation: ???The grandma and the grandpa that hit and sold the house drank water.
Embedding interpretation: * --- that hit the grandma that sold the house drank water.

*Interference-and-repair based analysis* predicts no such difference. According to it, parsing the incomplete sentence (17b) can also exhibit a missing-NP effect: parser thematically connects the ophaned verb $V_2$ to the head-NP by a repair mechnism, making all verbs linked to at least one argument. Therefore, similar results to Experiment 1 should be witnessed.

**5 Experiment 2**

Experiment 2 seeks to resolve unanswered questions from Experiment 1, specifically whether a missing-NP effect occurs in quasi-DCE sentences with reduced ambiguity and whether these sentences evoke a garden-path effect. Our *ambiguity-based analysis* predicts that quasi-DCE sentences like (17) should not trigger a missing-NP effect or a garden-path effect. Consequently, parsing the incomplete sentence should induce a P600 effect, indicative of syntactic violation, at the main clause verb, while no P600 effect, signifying structural reanalysis, should manifest at the second relativizer (relative to the first). In contrast, *interference-and-repair analysis* would predict that sentences like (17) should also elicit a missing-NP effect. As a result, no P600, which would indicate structural repair, and potentially an N400, denoting semantic incongruency, should appear at the main clause verbs.

**Materials**

80 quasi-DCE sentences with no local ambiguities were constructed and divided into complete and incomplete conditions as in (17). The verbs were so chosen that V2 and V1 take their NP-complements that have different animacy status. That is, V1 can only s-select an inanimate NP while V2 an animate one. Therefore, NPs at N1 and N2 position are invariably inanimate and animate, respectively. This should make implausible a conjunctive interpretation of RC verbs. The average frequency of words is 1.54 fpm, based on the corpora from BLCU Corpus Centre (BCC). The average frequency of first head-NP is 0.76 fpm and that of second head-NP is 0.71 fpm, with no significant difference between them after a two-sample *t*-test ($t(158) = .296$, $p = .767$).

Unlike materials in Experiment 1, we did not use non-human animals as animate NPs in Experiment 2's target sentences, in case that they may induce unwanted potentials. For instance, (18a) seems more semantic congruent than (18b):

(18) a. wéixiū wūdǐng de gōngrén
repair roof REL worker
"the worker who repairs the roof"
b. wéixiū wūdǐng de hóuzi
repair roof REL monkey
"the monkey who repairs the roof"

Note that we used non-human animals in Experiment 1 to reduce the possibility of a possessive interpretation of Mandarin *de* at the $RC_1$ region. This ambiguity should be not expected in $RC_1$ in the target sentences from Experiment 2, as the possessive interpretation of (19) is semantic incongruent:

(19) mài-le fángzi de nǎinai
sold house DE grandpa
RC interpretation: "the grandma who sold the house"
Possessive interpretation: * "*pro* sold the house's grandma"

In addition, to make sure that any possible significant effect at the matrix verb is purely caused by syntax (i.e., the incomplete condition lacking $RC_2$ head-NP), the nouns in the RC head-NP position were so chosen that they are a semantic meaningful pair, meaning that they can be coordinated congruently in semantics. For instance, in the VP "help the grandma and the grandpa" the two NPs combines congruently, whereas the two NPs in "help the grandma and the murderer" seems semantic incongruent and less expected.

All the fillers are adapted from Experiment 1, which all contain relative clauses.

**Participants**

Twenty-seven native speakers of Mandarin Chinese were recruited. All participants were given written informed consents and rewarded with extra credits when they completed the experiment (20 females, 7 males; age 18-25 ys, mean = 19.57, SD = 1.77). They all have normal or corrected-to-normal vision and did not report any neurological disorders or heart diseases. The experimental protocol was approved by the Ethics Committee of Soochow University, China. Two participants' datasets were excluded because they self-reported that they were physically and mentally uncomfortable to finish reading all the sentences. In addition, three datasets did not enter further analysis because of excessive noise. The artifact removal process excluded 12.89% of trials on average, from 22 participants included for further analysis. There was no significant difference in rejection rates among the conditions (Tukey correction, $F(5, 126) = .366$, $p = .871$).

**Procedure, EEG recording and statistical analysis**

All the procedures are the same as those in Experiment 1.

**Results and discussion**

*Priori mean time window (500-800 ms) looking for P600 effect*

For the main clause verbs comparison, the priori test found a significant positive cluster (cluster-mass t =162.9544, $p$ < .001, α = .05). This cluster is featured by a broad scalp distribution and spatial mass peak at electrode CP4.

At relativizers, the test found a marginally significant negative cluster, with its spatial mass peak over electrode PO7 (cluster-mass t = -32.7409, $p$ = .0503, α = .05). This cluster is more posteriorly distributed. Fig. 4 shows the grand-averaged waveform of PO7 and the topography of difference ERPs (second – first relativizer) where electrodes included in this cluster were highlighted. If reliable, this effect matches the spatial-temporal feature of *late posterior negativity* (LPN), which generally reflects task-related episodic memory retrieval (Mecklinger et al., 2016; Nie et al., 2013; Sommer et al., 2018). Since Mandarin relativizer *de* servers as the marker for the existence of a relative clause, processing it would trigger the retrieval of previously parsed argument structures (VPs and NPs). Therefore, the possible LPN effect at the second relativizer may imply that retrieving its argument structure requires more memory resources than that of the first relativizer.

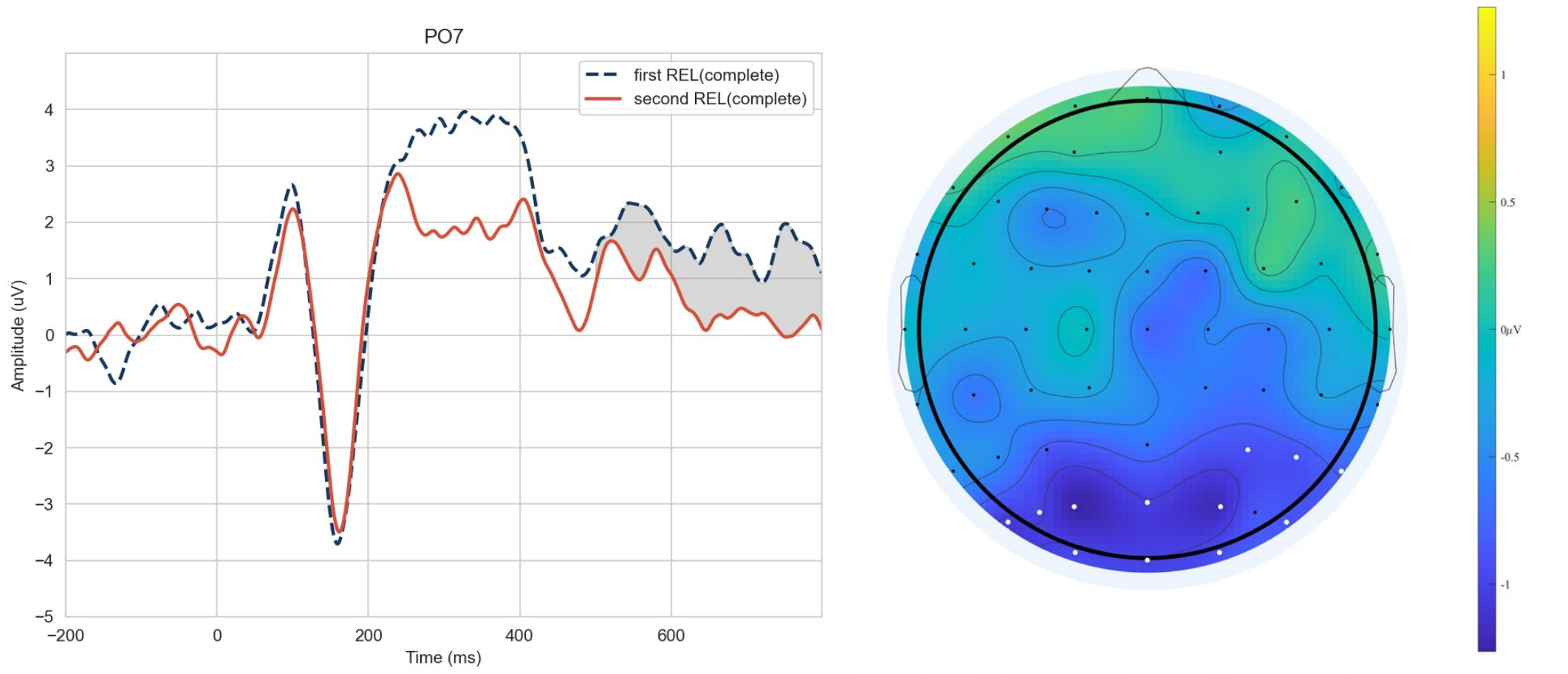

Fig. 4 Cluster mass results with a priori mean time window of 500-800 ms comparing brain potentials at relativizers. This graphic includes: a) grand-averaged waveforms of electrode PO7, where the effect is maximal; b) the topography of difference ERPs of the relativizers (second – first); electrodes that were included in this cluster were highlighted.

At head-NPs, the test found one significant positive cluster (cluster-mass t = 56.2568, $p$ = .02, α = .05). This cluster is featured by a fronto-central distribution with its spatial-mass peak at electrode FC4.

*Exploratory analysis (0-800 ms)*

Consistent with the priori time window test, the exploratory cluster mass test found one significant positive cluster at main clause verbs (cluster-mass t = 3606.4578, p < .001, α = .05), with its latency lasting from 510 to 800ms. This cluster was broadly-distributed over almost all the electrodes, with the greatest spatial-temporal mass effect over electrode CP4 around 640ms. Fig. 5 displays the grand-averaged waveform of CP4, the raster plot and the topography of difference ERPs (incomplete – complete structure) in a stepwise time interval of 100ms. The spatial-temporal distribution of this effect matches that of P600 (Hagoort et al., 1993; Hagoort & Brown, 2000; Osterhout & Holcomb, 1992; Swaab et al., 2011). We interpret this P600 as reflecting structural repair. This indicates that Mandarin native speakers experience no missing-NP effect in parsing quasi-DCE structure that has a reduced ambiguity at RC verbs. It gives a direct contrast to the result of Experiment 1 where missing-NP effect occurs. We can, therefore, rule out the prediction from *interference-and-repair* *analysis*.

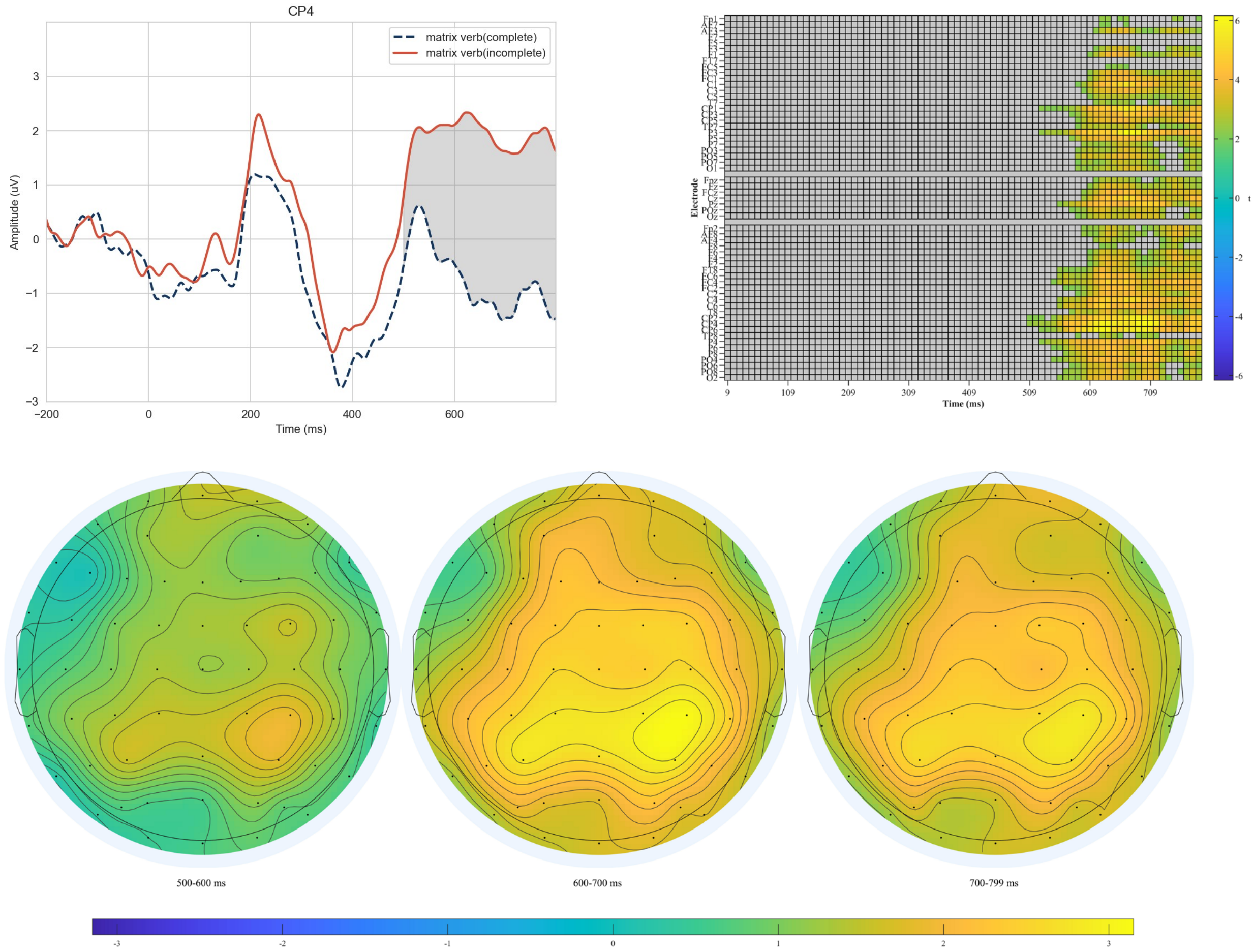

Fig. 5 Cluster mass results from main clause verbs comparison . This graphic includes: a) the grand-averaged waveforms of electrode CP4, where the effect is maximal; b) the raster plot showing the spatial-temporal distribution of this effect; and b) the topography of difference ERPs of matrix verbs (incomplete – complete structure) in a stepwise time interval of 100ms.

No significant clusters were reported at the comparison of the first and second relativizers (*ps* > .078). The possible LPN effect at the second relativizer, as reported by the priori mean time window test, suggests more processing load for the second relativizer than the first. In addition, the absence of a P600 at the second relativzer in Experiment 2 and the appearance of it in Experiment 1 suggests that the P600 effect in Experiment 1 reflects not an increase of integration cost but a garden-path effect—compared to the first relative marker, parsing the second relative marker in Experiment 1 requires structural reanalysis to satisfy relativization. Should it be an indication of an increase of integration cost, a similar effect would be expected in Experiment 2.

At the head-NPs, the permutation test found one significant positive cluster with a time window of 430-800ms, its spatial-temporal peak over electrode F4 around 700 ms (cluster-mass t = 1440.5211, *p* = .043, α = .05). Electrodes in the frontal region is more consistently involved than in the posterior region. Fig. 6 displays grand-averaged waveform of F4, the raster plot and the topography of difference ERPs (second – first head-NP) with a 100 ms stepwise time interval.

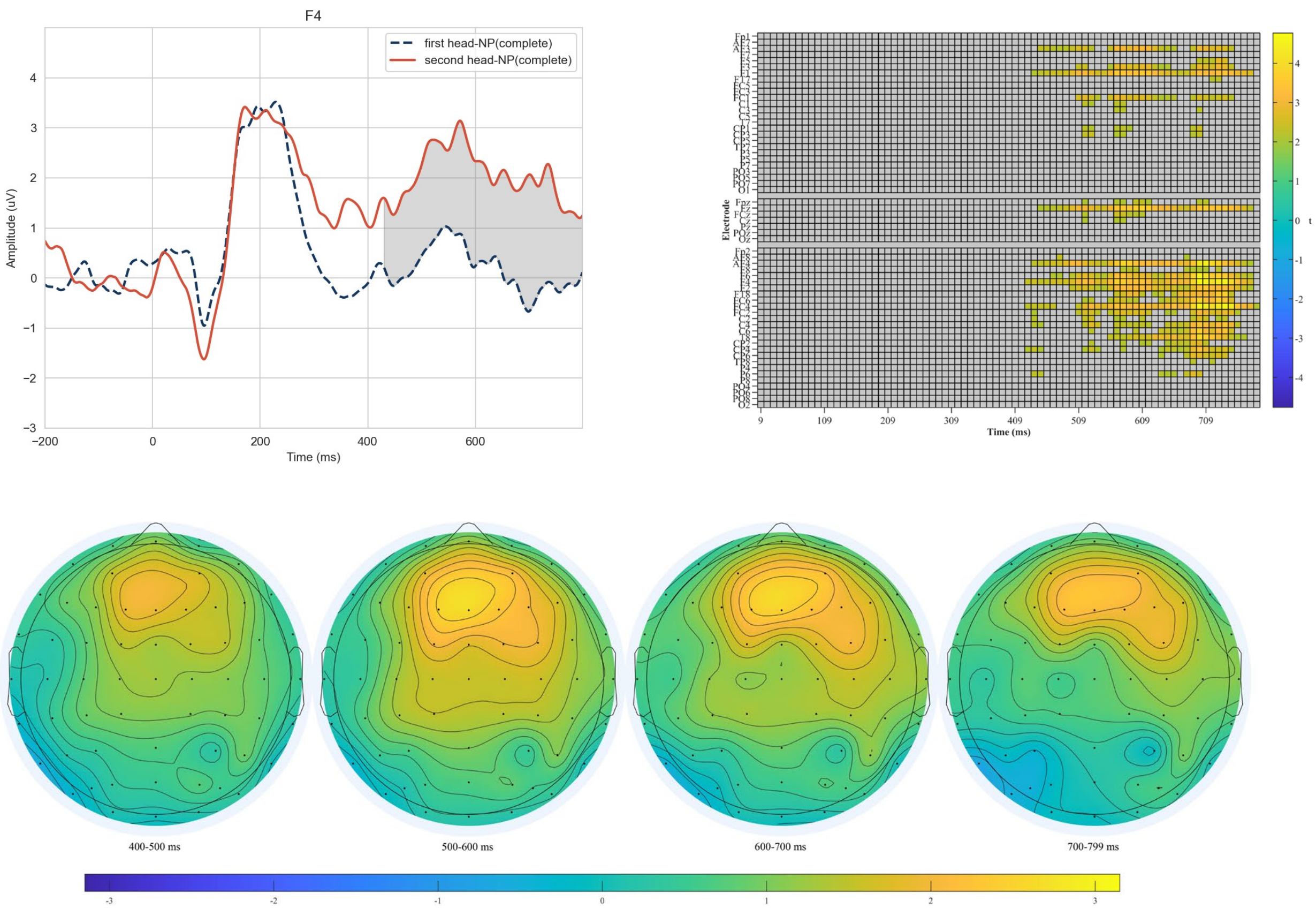

Fig. 6 Cluster mass exploratory results from head-NPs comparison. This graphic includes: a) the grand-averaged waveforms of electrode F4, where the effect is the strongest; b) the raster plot showing the spatial-temporal distribution of this effect; c) the topography of the difference wave (second – first head-NPs) with a stepwise time interval of 100ms.

The absence of P600 at second relativizer and the appearance of it at the second head-NP suggest that integration cost for relativization increases only at RC head-NPs for fulfilling filler-gap dependency. Since satisfying this dependency for the second RC head-NP trespasses $RC_1$ region (thereby longer distance and more interferences), processing the second head noun consumes more resources than the first one.

## 6 General discussion

To recap, Experiment 1 demonstrated that Mandarin native speakers exhibited a missing-NP effect when parsing a quasi-DCE sentence like (13), reiterated as (20). However, these results alone are insufficient to conclusively determine the driving force of the missing-NP effect. According to our *ambiguity-based analysis*, the parser initially conjoins RC verbs into a single, larger VP, which requires only one relativizer and one head-NP for relativization, thus leading to a missing-NP effect. Conversely, *interference-and-repair analysis* can attribute it to the complete thematic relation in the incomplete sentence, facilitated by a hypothesized repair mechanism. Regarding the centro-posterior P600 at the second relativizer, *ambiguity-based analysis* interprets it as reflecting structural reanalysis—the parser holds the conjunctive interpretation of RC verbs until the emergence of a second relativizer, leading to a garden-path effect; on the other hand, this could merely signify a heightened structural integration cost (Kaan et al., 2000).

(20) a. **Complete quasi-DCE sentence**

[matrix clause[RC2 _gap2 V2 dǎ-le [RC1 __gap1 V1 bāngzhù N1 nǎinai de filler1 & N2 yéyé]
[matrix clause[RC2 _gap2 V2 hit [RC1 __gap1 V1 helped N1 grandma REL filler1 & N2 grandpa]

de filler2 & N3 zuìfàn] matrix verb hē-le shuǐ]
REL filler2 & N3 the criminal] matrix verb drank water]

"The criminal that hit the grandpa that helped the grandma drank water."

b. **Incomplete quasi-DCE sentence**

[matrix clause[RC2 _gap2 V2 dǎ-le [RC1 __gap1 V1 bāngzhù N1 nǎinai de filler1 & N2 yéyé]
[matrix clause[RC2 _gap2 V2 hit [RC1 __gap1 V1 helped N1 grandma REL filler1 & N2 grandpa]
matrix verb hē-le shuǐ]
matrix verb drank water]

Conjunctive interpretation: ✔The grandpa that hit and helped the grandma drank water.
Embedding interpretation: * --- that hit the grandpa that helped the grandma drank water.

Experiment 2 offers more insight into the root cause of the missing-NP effect. Firstly, no missing-NP effect is observed for a quasi-DCE sentence (17), reiterated as (21), where a *semantic cue* is given in the RC verb region. This implies that reduced ambiguity in interpreting verbs leads the parser to reject a conjunctive interpretation of RC verbs, resulting in the grammatical violation of the incomplete sentence (21b), which, in turn, triggers a P600 effect associated with syntactic repair at the main clause verb. Secondly, the absence of a P600 effect for the second relativizer (relative to the first) infers that the centro-posterior P600 observed at the second relativizer in Experiment 1 is indicative of structural reanalysis rather than increases of integration cost.

(21) a. **Complete quasi-DCE sentence with reduced ambiguity**

[matrix clause[RC2 _gap2 V2 ōudǎ [RC1 __gap1 V1 mài-le N1 fángzi de filler1 & N2 nǎinai]
[matrix clause[RC2 _gap2 V2 hit [RC1 __gap1 V1 sold N1 the house REL filler1 & N2 the grandma]

de filler2 & N3 yéyé] main clause verb hē-le shuǐ]
REL filler2 & N3 the grandpa] main clause verb drank water]

"The grandpa that hit the grandma that sold the house drank water."

b. **Incomplete quasi-DCE sentence with reduced ambiguity**

[matrix clause[RC2 _gap2 V2 ōudǎ [RC1 __gap1 V1 mài-le N1 fángzi de filler1 & N2 nǎinai]
[matrix clause[RC2 _gap2 V2 hit [RC1 __gap1 V1 sold N1 the house REL filler1 & N2 the grandma]
he filler2 & N3 yéyé] ---] main clause verb hē-le shuǐ]
and filler2 & N3 the grandpa] ---] main clause verb drank water]

Conjunctive interpretation: ???The grandma and the grandpa that hit and sold the house drank water.
Embedding interpretation: * --- that hit the grandma that sold the house drank water.

The sole structural difference between (20) and (21) lies in the presence of a semantic cue. Disregarding ambiguity, the interference-and-repair analysis would predict analogous outcomes for Experiments 1 and 2 regarding both the manifestation of a missing-NP effect and the processing dynamics at the second relativizer. The contrasting results from Experiment 1&2 suggest that: i) ambiguity at RC verb region induces a missing-NP effect in (20b) and a garden-path effect in (20a); ii) a semantic cue can mitigate this ambiguity, thus eliminating the missing-NP effect in (21b) and the garden-path effect in (21a).

Alternatively, it may be contended that the disparate outcomes of Experiments 1 and 2 are attributable to the reversibility of thematic relations. In Experiment 1, target sentences comprise three animate NPs, with each pair of thematic relations being reversible (allowing for interchangeable assignment of Agent/Patient theta-roles). In contrast, Experiment 2 involves two animate NPs and one inanimate NP, with only a single pair of thematic relations being reversible. The difference in NPs' animacy status may well contribute to this disparate result, as many studies have found that animacy plays a role in processing asymmetry of relative clauses (DeDe, 2015; He & Chen, 2013; Lowder & Gordon, 2012; Mak et al., 2006; Mak et al., 2002; Perera & Srivastava, 2016; Traxler et al., 2005; Wu et al., 2011; Wu et al., 2012). However, prior research has evidenced the missing-VP effect in both thematic-reversible (Christiansen & MacDonald, 2009; Frank et al., 2021; Frank et al., 2016; Vasishth et al., 2010) and -irreversible DCE sentences (Frank & Ernst, 2019; Gibson & Thomas, 1999). Hence, while reversibility may influence the processing of (quasi) DCE structures as indicated by (Gibson et al., 2013), it's unlikely to be the crucial determinant of a missing-VP/NP illusion.

So far, our discussion has been confined to the Mandarin quasi-DCE structure. Recall that a quasi-DCE structure is constructed by relocating the self-embedding RC segment of the DCE structure into the subject position of the main clause. In other words, the sole distinctions between the two structures lie in the location of RC and the arrangement of their sequences, with the quasi-DCE structure featuring two adjacent verbs as in "VVN*de*N*de*NVN", contrasted with the DCE structure's three in "NVVVN*de*N*de*N". Via this transformation, the complexity of it is much more reduced compared to that of DCE structure. As deliberated in Section 3, if Mandarin quasi-DCE structure exhibit a garden-path effect due to ambiguity in the RC verb region, then it is implausible not to observe a similar effect affecting the parsing process of DCE structure, whose complexity is only higher. Given the fact that no existing theories adequately explain the missing-NP effect in both the ideally unambiguous Mandarin DCE structure and quasi-DCE structure, ambiguity emerges as the most viable factor behind the missing-NP effect in Mandarin DCE structure. In extreme cases, the effect of ambiguity could be so pronounced that upon receiving an unfinished sequence like "NVVVN" such as (22), it may recognize it as a grammatically complete sentence, despite its semantic incongruity.

(22) Jǐngchá dàibǔ ānwèi mài-le nǎinai…
The police [$_{V1}$ arrested] [$_{V2}$ consoled] [$_{V3}$ sold] [$_{N1}$ grandma]
"The police arrested, consoled and abused the grandma."

In addition, the processing constraints of Mandarin DCE structure can also be attributed to ambiguity. As discussed in Section 3, building an embedding interpretation for the verbs demands a greater sequence of operations (i.e., memory load) than building a conjunctive interpretation. This suggests that the parser may retain a conjunctive interpretation for the verbs, until the appearance of a disambiguating relative marker *de*, as evidenced by the garden-path effect in Experiment 1. Therefore, in extreme cases parsing this structure may trigger a garden-path effect at every occurrence of a relativizer, each adding extra processing loads. Table 1 displays plausible tree diagrams of an incremental parsing process before and at the disambiguating relativizers.

This proposal does not negate the impact of memory limitations or interference in the processing constraints of the Mandarin DCE structure. In an ideally unambiguous Mandarin DCE structure, parsing each verb that is not the main clause verb triggers the prediction of a relative clause (embedding interpretation). In that case, the appearance of a relativizer *de* should not lead to structural reanalysis. And according to *resource-based account* and *interference-based account*, the highest memory load for parsing should be witnessed at the second head-NP, at which juncture native Mandarin speakers may struggle with resolving its long-distance nested dependencies.

But as we have discussed in Section 2.2, theories that exclusively focus on memory constraints and do not take ambiguity into consideration fall short of explaining the missing-NP effect in Mandarin DCE structure. Hence, we align with *experience-based account* perspective positing that grammaticality illusion in parsing DCE structure is language-dependent, contingent on the distinct word-order of that language, and therefore, not a universal phenomenon. Meanwhile we suggest that the parser's limitation in processing structures with multiple levels of center-embedding should be a universal property.

Table 1 the tree diagrams of parsing Mandarin DCE sentence. For simplicity, words that indicate noun phrase and verb phrase are abbreviated by "N" and "V", where "$N_p$" means not encountered but predicted noun phrase.

| before first *de* | at first *de* | at second *de* |
|---|---|---|
| 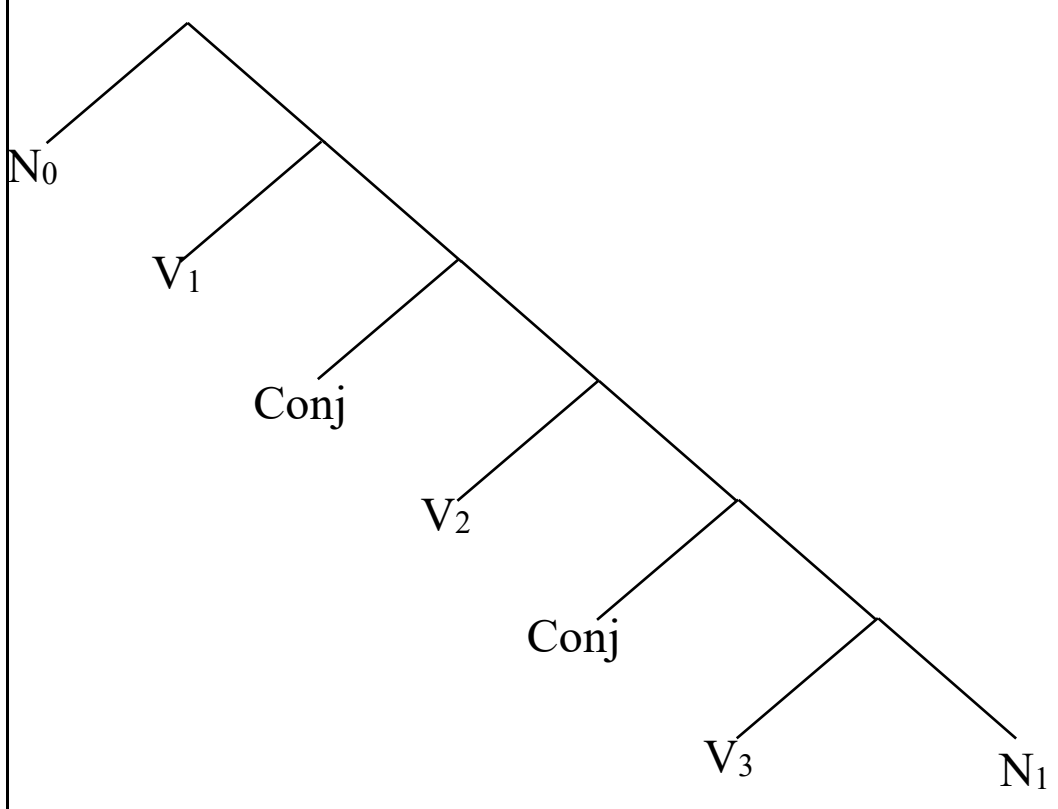  | 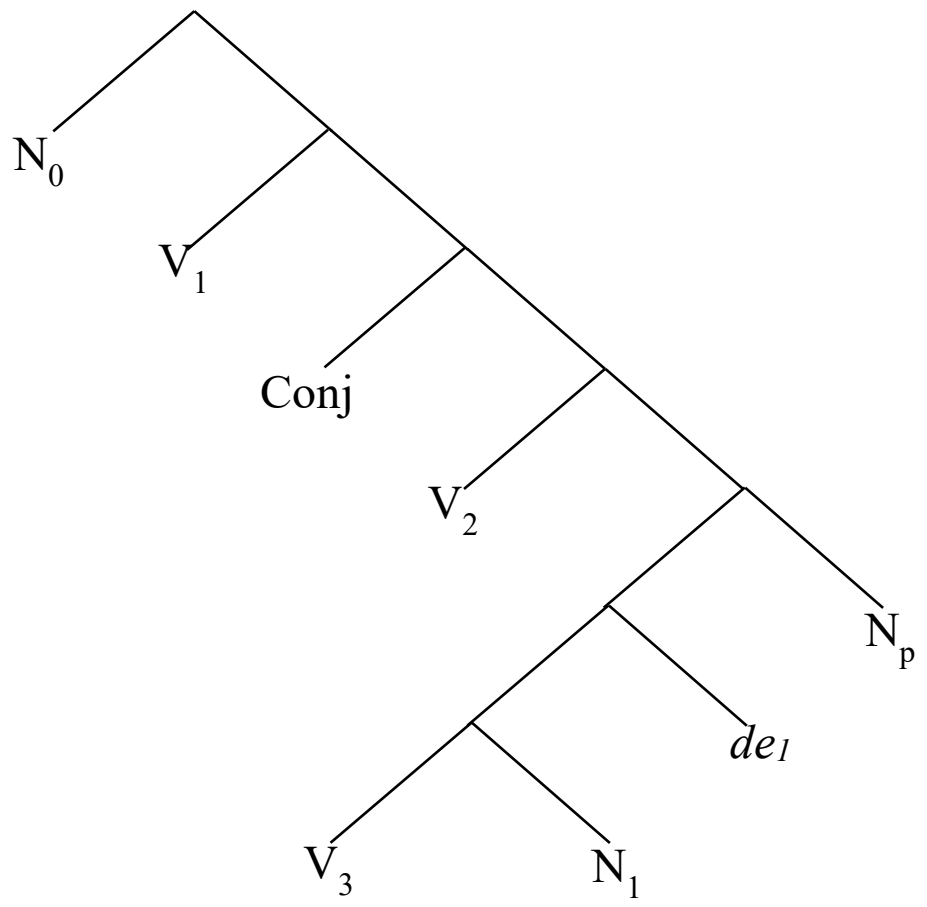  | 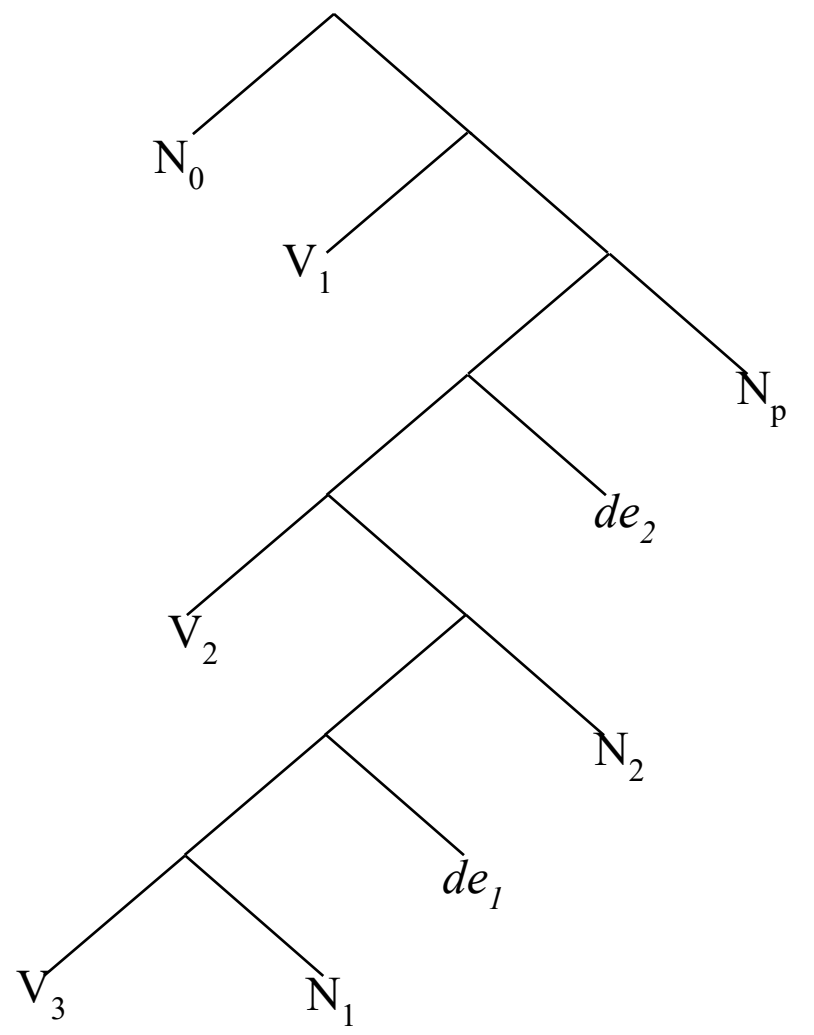  |

Finally, let's come back to the persistent question from Section 3; that is, whether there are ways to mitigate this ambiguity. Because of Mandarin RC's head-final feature, it is not plausible to completely resolve the ambiguity in interpreting verbs. However, there might be some ways to reduce this ambiguity, to the extent that the parser might not initiate a conjunctive interpretation for the verbs. For instance, in Experiment 2, we utilized a *semantic cue* for the parser to see the impossibility of a conjunctive interpretation for the RC verbs in quasi-DCE structure. To adapt this approach for Mandarin DCE structure, we could construct V2 as taking an inanimate NP complement while V1 and V3 as taking animate NP complements, shown in (23). This should bar a conjunctive interpretation of the verbs, so that missing-NP effect would not occur, corroborating the findings from Experiment 2.

(23) Gōngrén [$_{V3}$ chāichú] [$_{RC2\ _gap2}$ $_{V2}$ zhùzhe [$_{RC1\ __gap1}$ $_{V1}$ bāngzhù $_{N1}$ sūnzi
The worker [$_{V3}$ demolished] [$_{RC2\ _gap2}$ $_{V2}$ sheltered [$_{RC1\ __gap1}$ $_{V1}$ helped $_{N1}$ the grandson
de $_{filler1\ \&\ N2}$ nainai] ---]
REL $_{filler1\ \&\ N2}$ grandma] —]
Conjunctive interpretation: ???The worker demolished (and) sheltered the grandma who helped the grandson.
Embedding interpretation: *The worker [demolished [--- sheltered]] the grandma who helped the grandson.

However, the efficacy of this semantic cue was only confirmed for quasi-DCE structures. Given the increased complexity of Mandarin DCE structures compared to quasi-DCE structures, the semantic cue might be ineffective due to the heightened memory demands. The parser may ignore this cue and persist with a conjunctive interpretation of the verbs, despite resulting in significant semantic inconsistencies. Since this study cannot conclusively address this issue, it remains an open question for future research.

Studies that focused on SRC/ORC asymmetry of Mandarin RC have identified some ways to signal the presence of a relative clause (see Jäger et al., 2015; Lin & Bever, 2011, for details). We discuss one approach that is relevant to the ambiguity we have been facing.

In (23a) the main clause verb is available to conjoin the RC verb, therefore creating a local ambiguity. Inserting a classifier *nàgè* (a NP modifier that in some cases serves as a cue for an upcoming RC) in between $V_1$ and $V_2$ bars a conjunctive interpretation.

(23) a. Jingcha dàibǔ$_{V1}$ bangzhu$_{V2}$ nainai$_{N1}$ …
The police arrested$_{V1}$ helped$_{V2}$ the grandma$_{N1}$ …

Conjunctive interpretation: "The police arrested (and) helped the grandma …"
Embedding interpretation: "The police arrested (someone who) helped the grandma …"

b. Jingcha dàibǔ$_{V1}$ nàgè bangzhu$_{V2}$ nainai$_{N1}$ …
The police arrested$_{V1}$ CL helped$_{V2}$ the grandma$_{N1}$ …

Conjunctive interpretation: impossible
Embedding interpretation: "The police saw (someone who) helped the grandma …"

This approach works fine for a single relative clause. But when it comes to a DCE structure, the situation is much more complicated. First, the classifier *nàgè* can not only serve as a cue for an upcoming RC, but in some cases can be a nominal head that equals to "that" or "this" in English. For example, (24) is perfectly comprehensible for a Mandarin native speaker:

(24) Wǒ xiǎngyào zhège, dàn bùyào nàgè.
I want this, but not want that.

"I want this, but not that."

Second, in a sentence where two NPs modified by their classifiers have exactly same lexical item, one of the nouns can be phonetically dropped, creating a representation like (25):

(25) Lǎobǎn kuāzàn-le zhège dànshì zhǐzé-le nàgè yuángōng
The minister praised this but blamed that employee

"The minister praised this (employee) and promoted that employee."

These two properties of the classifier *nàgè* together adds two ambiguities when it is inserted in a Mandarin DCE structure as in (26). The first ambiguity allows the parser to build a conjunctive interpretation "The police arrested that (person) and assaulted this (person)", while the second ambiguity makes the sentence interpreted as "The police arrested that (man) and assaulted this man who sold the house", which is structurally complete.

(26) Jǐngchá$_{SUB}$ [[dàibǔ$_{V1}$ nàgè ōudǎ$_{V2}$ zhège]$_{ambiguity1}$ mài-le$_{V3}$ fángzi$_{N1}$
The police arrested CL assaulted CL sold house

de nánrén$_{N2}$]$_{ambiguity2}$ …
REL man …

Desired meaning: The police arrested (someone) who assaulted the man who sold the house.

Adding new ambiguities may make the parsing process even harder, as it could involve more structural reanalyses. Therefore, this approach may not be viable for resolving this ambiguity, and requires empirical evidence from future studies.

**Conclusion**

Our findings indicate that the less complex Mandarin quasi-DCE structure, with its complexity reduced by positioning self-embedding RCs in the subject position, exhibits a "missing-NP effect" similar to Huang and Phillips (2021) when an RC head-noun and relativizer are omitted. However, this effect vanished when we provided the parser with a semantic cue, which bars the RC verbs from a conjunctive interpretation. We argued that the root cause of this disparity is the inherent ambiguity in interpreting Mandarin verbs. The most efficient way of constructing an unfinished sequence "VVNN*de*N…" into a hierarchical structure is to adopt a conjunctive interpretation of the verbs. In the case of Mandarin quasi-DCE structure, this interpretation will make an incomplete quasi-DCE sentence grammatical, thereby creating an illusion of a missing-NP effect. When the ambiguity is reduced, no such effect will occur due to the impossibility of the conjunctive interpretation of the verbs. We noted that if not taking into consideration the ambiguity in interpreting Mandarin verbs, none of the existing theories can provide a sound explanation for these findings. This suggested that ambiguity should also be the cause for the missing-NP effect and the processing limitation in Mandarin DCE structure, which is more complex and has more adjoint verbs than Mandarin quasi-DCE structure. In that case, the so-called grammaticality illusion in the missing-NP Mandarin DCE structure is instead not "illusion" at all, since the conjunctive interpretation of it is perfectly grammatical. This may explain why current language models like *lossy-context surprisal model* failed to predict a missing-NP effect, as in an ideally unambiguous DCE structure no verbs would attain a conjunctive interpretation.